\documentclass{article}
\pdfoutput=1

    \usepackage[preprint, nonatbib]{neurips_2020}

\usepackage[utf8]{inputenc} 
\usepackage[T1]{fontenc}    
\usepackage{url}            
\usepackage{booktabs}       
\usepackage{amsfonts}       
\usepackage{nicefrac}       
\usepackage{microtype}      

\usepackage{hyperref}       
\usepackage{xcolor}
\hypersetup{
  hidelinks,
  colorlinks,
  linkcolor={red!50!black},
  citecolor={blue!50!black},
  urlcolor={blue!80!black}
}
\usepackage{graphicx}
\usepackage{amsmath}
\usepackage[titlenumbered,ruled,linesnumbered]{algorithm2e}
\usepackage{floatrow}
\usepackage{xspace}
\newcommand{\bi}{\begin{itemize}}
\newcommand{\ei}{\end{itemize}}

\makeatletter
\DeclareRobustCommand{\cev}[1]{%
  {\mathpalette\do@cev{#1}}%
}
\newcommand{\do@cev}[2]{%
  \vbox{\offinterlineskip
    \sbox\z@{$\m@th#1 x$}%
    \ialign{##\cr
      \hidewidth\reflectbox{$\m@th#1\vec{}\mkern4mu$}\hidewidth\cr
      \noalign{\kern-\ht\z@}
      $\m@th#1#2$\cr
    }%
  }%
}
\makeatother

\definecolor{gray}{rgb}{.5,.5,.5}

\makeatletter
\DeclareRobustCommand\onedot{\futurelet\@let@token\@onedot}
\def\@onedot{\ifx\@let@token.\else.\null\fi\xspace}

\def\ie{\emph{i.e}\onedot}

\SetKwIF{If}{ElseIf}{Else}{if}{}{else if}{else}{end if}%

\title{Efficient Data Association and Uncertainty \\Quantification for
Multi-Object Tracking}

\author{%
  David S.~Hayden, Sue Zheng, John W.~Fisher III \\
  Computer Science and Artificial Intelligence Laboratory \\
  Massachusetts Institute of Technology \\
  Cambridge, MA 02142 \\
  \{dshayden, szheng, fisher\}@csail.mit.edu
}

\begin{document}
\maketitle

\begin{abstract}
  Robust data association is critical for analysis of long-term motion
  trajectories in complex scenes. In its absence, trajectory precision suffers
  due to periods of kinematic ambiguity degrading the quality of follow-on
  analysis. Common optimization-based approaches often neglect uncertainty
  quantification arising from these events. Consequently, we propose the Joint
  Posterior Tracker (JPT), a Bayesian multi-object tracking algorithm that
  robustly reasons over the posterior of associations and trajectories. Novel,
  permutation-based proposals are crafted for exploration of posterior modes
  that correspond to plausible association hypotheses. JPT exhibits more
  accurate uncertainty representation of data associations with superior
  performance on standard metrics when compared to existing baselines. We also
  show the utility of JPT applied to automatic scheduling of user-in-the-loop
  annotations for improved trajectory quality.
\end{abstract}


%
\section{Introduction}
\label{sec:intro}
\begin{figure}[t]
  \centering
  \includegraphics[width=\textwidth]{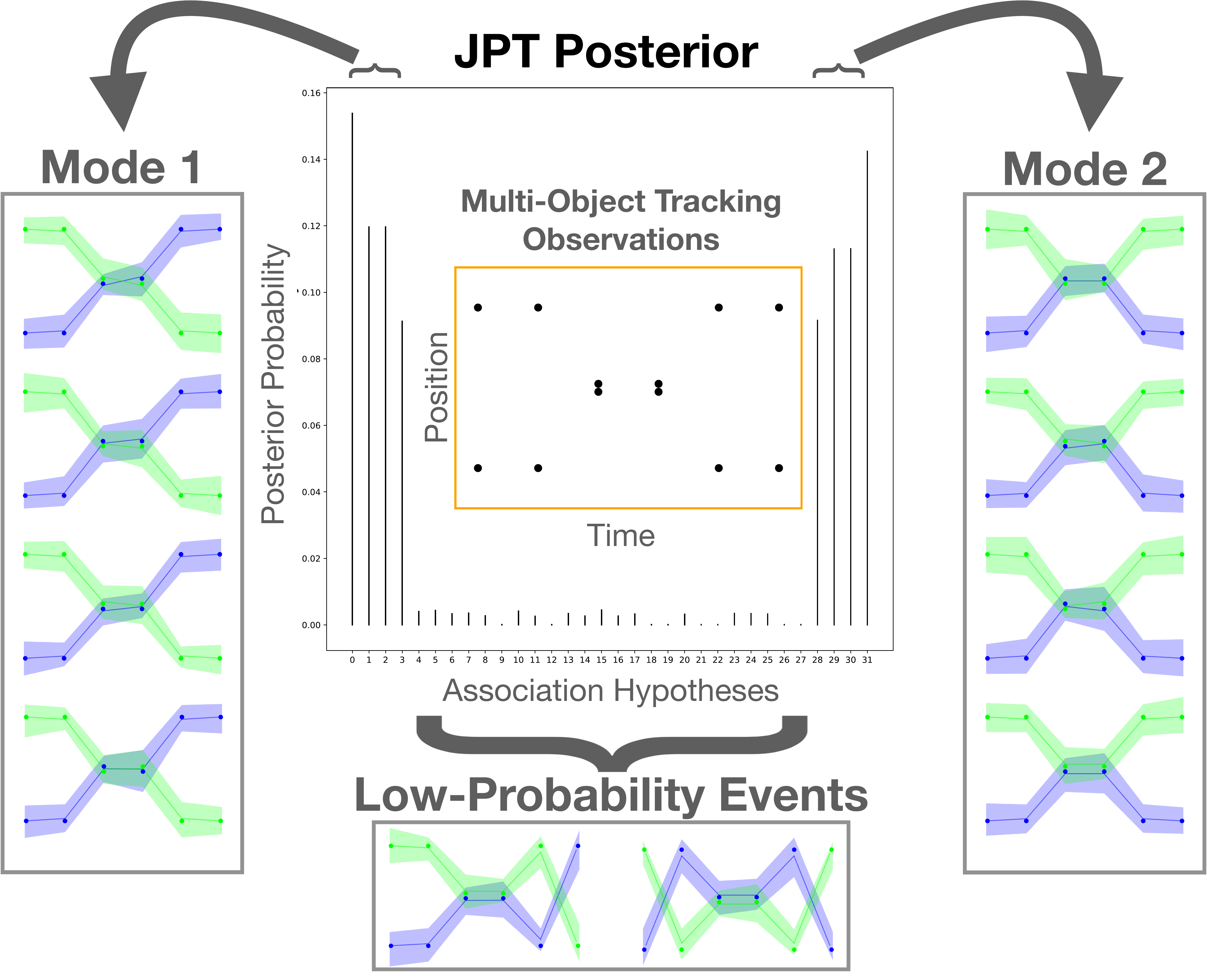}
  \caption{Multi-object tracking observations over time (orange box, middle)
  where two objects (green, blue) begin separated--briefly converge--then
  diverge. Such ambiguities are represented as distinct modes (left, right)
  which the Joint Posterior Tracker excels at exploring while avoiding
  low-probability events (bottom).
  }
  \label{fig:teaser}
\end{figure}
In multi-object tracking the trajectories of an unknown number of objects are
estimated from noisy observations over time. Assigning observations to objects
is known as the \emph{data association problem}. The complexity of the
multidimensional assignment formulation is factorial in the
number of observations at each time and exponential in the number of
timesteps making it NP-hard \cite{blackman1999design}.

\emph{All} approaches to multi-object tracking reason over data
associations, but \emph{few} represent uncertainty explicitly, much
less make it available for subsequent tasks. Traditional applications
in security \cite{benfold2011stable}, surveillance
\cite{oh2011large}, sensor networks \cite{sastry2006instrumenting}
and robotic localization \cite{nieto2003real} favor real-time
performance. More recently, there is increased interest in the use tracking for
follow-on decision making and analysis including the creation of
gold-standard datasets \cite{Vondrick2013}, sports
analytics \cite{Ferryman2020}, study of animal
behavior \cite{zhou2019atypical} and cellular dynamics
\cite{arasteh2018multi}. Such applications benefit significantly from \emph{accurate}
representations of uncertainty to inform subsequent analysis.

%
%

In response, we develop the Joint Posterior Tracker (JPT), a
Bayesian multi-object tracker that emphasizes \emph{joint} uncertainty
over association hypotheses and object trajectories.
We present model details in Section~\ref{sec:model}. In Section
\ref{sec:inference} we develop a novel Metropolis-Hastings inference
procedure that, under some model parameterizations, generalizes the
Extended-HMM proposals of \cite{neal2004inferring}. JPT inference is
exact, efficient and not limited by gating heuristics. In
Section~\ref{sec:experiments}, we show that JPT explores posterior
modes much more completely and efficiently with superior performance
on standard multi-object tracking metrics on scientific and sports
datasets as compared to a standard baseline. Finally, we show that JPT
enables the automatic scheduling of a small number of disambiguations
that facilitate rapid improvement in trajectory quality.

Figure~\ref{fig:teaser} illustrates JPT reasoning over multimodal uncertainty. Two objects
begin well separated, become kinematically ambiguous, then
separate. Absent additional information, it is unclear whether or not
their paths crossed. Evaluation of
association events shows that the resulting JPT posterior accurately
captures trajectory uncertainty via two modes--one for crossing, one for not. \emph{Within} each
mode, associations switch when objects are close but inferred trajectories
remain similar whereas \emph{between} modes, associations switch in ways that
dramatically impact the inferred trajectories (and hence subsequent analysis).
%
%

%
\paragraph{Related Work}
\label{sec:related}
Approaches to multi-object tracking can be distinguished in several ways.
First, whether they process measurements one frame at a time (single-scan,
\cite{segal2013latent, khan2005mcmc, breitenstein2010online}), multiple frames
at a time (multi-scan, \cite{kim2015multiple}) or all at once (batch,
\cite{butt2013multi, oh2009markov, wang2014tracklet}). Second, whether they construct point estimates (as
in optimization), \cite{kim2015multiple,
hamid2015joint, wang2014tracklet} or
entertain multiple solutions (sampling or variational methods,
\cite{segal2013latent, turner2014complete}). Third, whether or not they employ
gating heuristics that restrict possible hypotheses \cite{kim2015multiple,
oh2009markov}.  JPT is a batch, sampling-based tracker with no gating
heuristics that reasons over the joint distribution of an unknown number of
objects, their trajectories and the association of objects to observations.

Many recent approaches to multi-object tracking focus on sophisticated
appearance, motion or shape modeling in an optimization-based
framework \cite{hamid2015joint, leibe2007coupled, kim2015multiple,
  zamir2012gmcp, butt2013multi}. While providing a single
point-estimate, they forego representing uncertainty in assignments,
and do not permit recovery from errors. Regardless of the quality of
the appearance model, errors are certain to occur in complex scenes.


Monte-Carlo approaches represent uncertainty via sampled realizations.
These include \cite{segal2013latent}, \cite{khan2005mcmc},
\cite{breitenstein2010online}, but each are single-scan, filtering-based
approaches, that do not incorporate future information.
\begin{table}
  \centering
  \large
  \begin{tabular}{|c| c c c c|}
    \hline
                        & \textbf{JPT}   & MCMCDA \cite{oh2009markov}    & Var \cite{turner2014complete}       & BP \cite{williams2010data} \\
    \hline
    Uncertainty         & \checkmark  & \checkmark & \checkmark & \checkmark \\
    Exact Posterior     & \checkmark  & \checkmark &            &  \\
    General MOT         & \checkmark  & \checkmark & \checkmark &  \\
    \hline
  \end{tabular}
  \caption{Multi-Object Trackers capable of quantifying uncertainty in data
  association. 
  JPT (Joint Posterior Tracker): our method; 
  MCMCDA (Markov Chain Monte Carlo Data Association);
  Var (Variational Tracker);
  BP (Belief Propagation Tracker). 
  }
  \label{table:related}
\end{table}
Table~\ref{table:related} summarizes related works that represent some
degree of uncertainty, such as marginal uncertainty using belief
propagation \cite{williams2010data} and approximate uncertainty using
variational methods \cite{turner2014complete}. The former treats a
fixed number of objects, while the latter samples from a
variational approximation rather than the true posterior.

Markov-Chain Monte Carlo Data Association (MCMCDA) and its variants
\cite{oh2009markov, benfold2011stable, ge2008multi} is most closely
related to JPT. MCMCDA employs sampling and can be run online or in
batch. It represents posterior uncertainty in data association, but
does not incorporate information from the future (it performs
filtering as opposed to smoothing or joint inference, even when
running in batch). Furthermore, MCMCDA precomputes data structures
that heavily rely on gating heuristics. This precludes uncertainty
representation of some association hypotheses.
Nevertheless, it is the only tracker the authors are aware of that
can, in principle, represent posterior uncertainty, samples from its
posterior exactly (modulo gating heuristics) and treats the general
multi-object tracking problem. As such, we use batch MCMCDA as a
baseline for comparison to JPT.


%
\section{Bayesian Multi-Object Tracking}
\label{sec:model}
The multi-object tracking problem is to partition a set of observations across
time into collections of objects such that every observation must be assigned
and no two objects can claim the same observation. In its most general
formulation, there can be clutter (false-positives), missing detections, unknown
number of objects, and arbitrary object arrival and departure times.
Multi-object tracking can be formulated in several ways; we discuss this further
in Appendix~\ref{sec:maf} and define the multidimensional assignment formulation as it
is most closely related to JPT.

\begin{figure}[]
  \centering
  \includegraphics[width=0.9\textwidth]{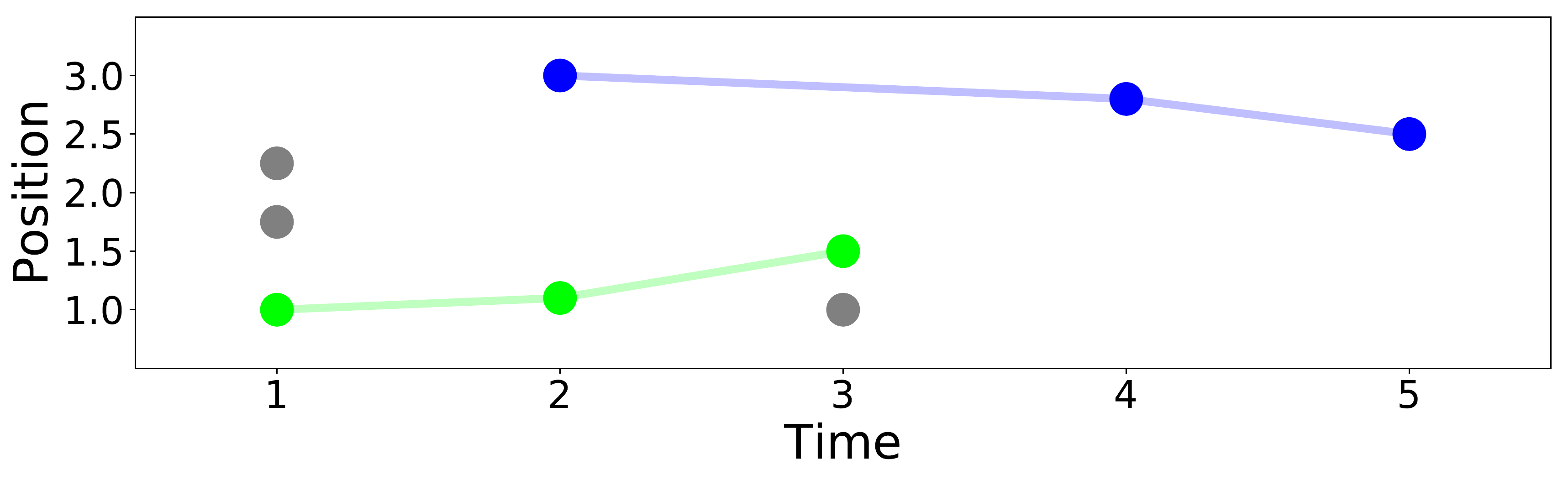}
  \includegraphics[width=0.9\textwidth]{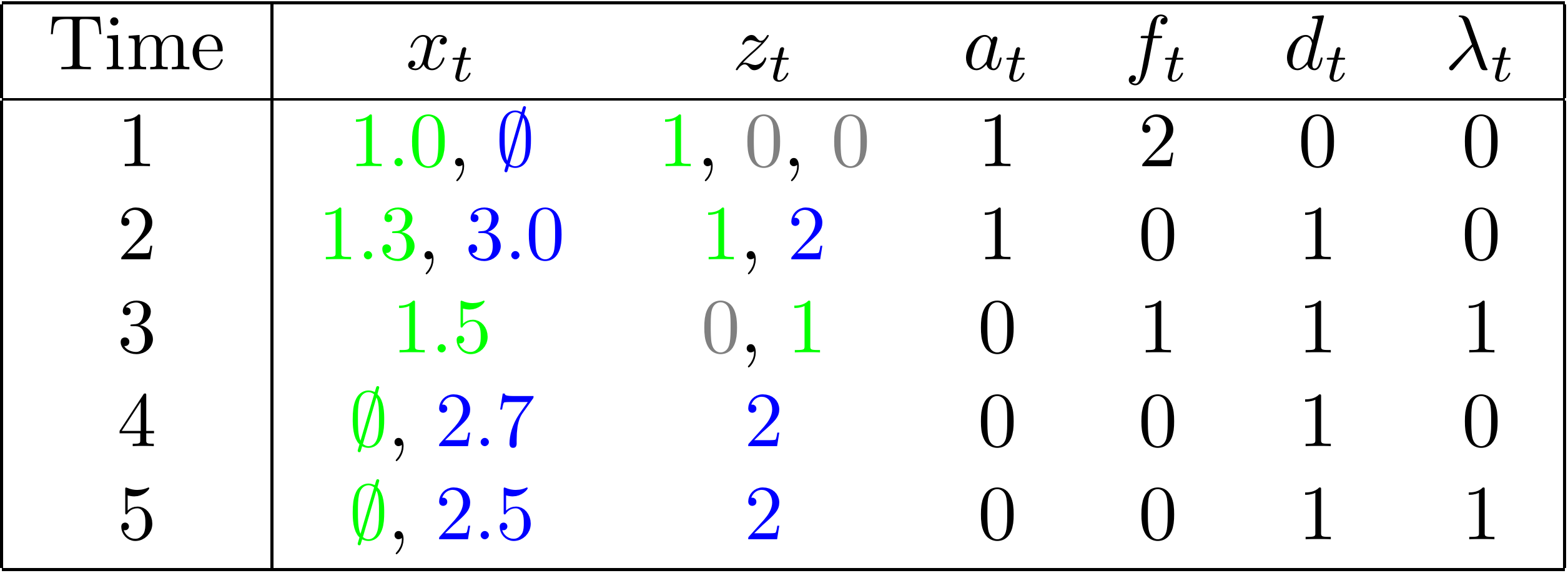}
  \caption{JPT's Latent Representation. (\textbf{Top}): Observations
  $y$ (circles) are colored by their association (green/blue for
  either of two objects, gray for clutter) and connected by sampled
  trajectories. (\textbf{Bottom}): Trajectories $x$, associations $z$ and
  counts $M = \{ a_t, f_t, d_t, \lambda_t \}_{t=1}^T$. Objects are observed at distinct arrival and departure times.
  Trajectories are length-$T$, padded by $\emptyset$ before arrival and after
  departure. We marginalize over states with missing detections (e.g. blue at
  $t=3$).
  }
  \label{fig:example_mot}
\end{figure}
JPT defines a joint distribution on trajectories and assignments whereas
MCMCDA defines a posterior on assignments alone. Both operate in batch,
but a key difference is that MCMCDA \emph{only} considers past information when
estimating trajectories via filtering. In contrast, JPT samples from a
joint distribution over associations and trajectories that accounts for past and
future information. Hence, there is a distribution over trajectories for any set
of associations.

To reason over a joint distribution on trajectories and associations, we must
define a generative model for observations $y = \{ y_1, \ldots, y_T \}$ over all
times $1, \ldots, T$ where $y_t = \{y_{tn}\}_{n=1}^{N_t}$. Vector-valued
observation $y_{tn}$ is the $n^{\text{th}}$ observation at time $t$ and has
dimension $D_y$. $N_t$ is the total number of observations at time $t$.

Associations $z$ define a partitioning of $y$ into objects and clutter and
trajectories $x$ are the latent states of objects over
all times. For clarity of exposition since our goal is accurate representation
of posterior uncertainty, we only model object locations. One can include a
shape or appearance model without modification of \emph{any} equation in this work.
Next, we define the latent representation and generative model for JPT.
Throughout, Figure~\ref{fig:example_mot} can be used to ground definitions in a
toy example.

\subsection{Event Counts $p(M)$}
JPT explicitly models clutter (false-positive) and missing detections, as well
as arbitrary arrival and departure times for an unknown number of objects.  At
each time $t$, counts of new object arrivals $a_t$, clutter observations $f_t$,
existing object detections $d_t$ and existing object departures $\lambda_t$ are
modeled as
\begin{equation}
  \begin{aligned}
    a_t \sim \text{Pois}(a_t \mid \lambda_b) \mkern18mu 
    f_t \sim \text{Pois}(f_t \mid \lambda_f) \mkern18mu 
    d_t \sim \text{Bin}(d_t \mid e_{t-1}, p_d) \mkern18mu 
    \lambda_t \sim \text{Bin}(\lambda_t \mid d_t, p_{\lambda}) \mkern18mu
  \end{aligned}
  \label{eqn:counts}
\end{equation}
where $e_0 = d_0 = 0$ and $e_t = e_{t-1} + a_t - \lambda_t$ are counts of
existing objects (those that arrived at some time $t' \leq t$ and have not
yet departed). Prior parameters $\lambda_b, \lambda_f$ are the new object arrival and false alarm rates and $p_d, p_{\lambda}$ are the detection and departure probabilities for existing objects.
Every object is assumed to be observed at least twice: when it arrives and when it departs. Denote the set of all event counts as
$M = \{ M_1, \ldots, M_T \}$ where $M_t = \{ a_t, f_t, d_t, \lambda_t \}$. From
Eqn.~\ref{eqn:counts}, the generative model for latent counts $M$ is,
\begin{equation}
  \begin{aligned}
    p(M) &= \prod_{t=1}^T p(M_t \mid M_{t-1}) 
    &= \prod_{t=1}^T p(a_t)\ p(f_t)\ p(d_t \mid e_{t-1})\ p(\lambda_t \mid d_t).
  \label{eqn:M}
  \end{aligned}
\end{equation}
\subsection{Associations $p(z \mid M)$}
JPT represents the association of each observation to an
integer-labeled object or clutter. Let the association of observation
$y_{tn}$ be the latent random variable $z_{tn} \in \mathbb{Z}^{+} \cup \{ 0 \}$
where $z_{tn} = k > 0$ if $y_{tn}$ is associated to target $k$ at time $t$ and
$z_{tn} = 0$ if $y_{tn}$ is associated to clutter. Define an association
hypothesis as the set of all associations $z = \{ z_1, \ldots, z_T \}$ for $z_t
= \{ z_{tn} \}_{n=1}^{N_t}$.

Conditioned on event counts $M$, association hypotheses $z$ have a uniform prior
over the space of possible associations subject to constraints that enforce that
all observations are either associated to an object or clutter (satisfied by
definition of $z_{tn}$), that an object claim at most one observation at each
time $t$ (first constraint in Eqn.~\ref{eqn:z}) and that associations be
consistent with event counts $M$ (remaining constraints in Eqn.~\ref{eqn:z}):
\begin{equation}
  \begin{aligned}
    & p(z \mid M) \propto 1 \text{ if } 
    &\begin{cases}
    | \{ n : z_{tn} = k \} | \leq 1 \quad \forall k>0, \forall t \\
    f_t = | \{ n : z_{tn} = 0 \} | \quad \forall t \\
    a_t = | \{ k > 0 : z_{tn} = k \text{ and } z_{t' n} \neq k \text{ for all } t' < t \} | \quad \forall t \\
    \lambda_t = | \{ k > 0 : z_{tn} = k \text{ and } z_{t' n} \neq k \text{ for all } t' > t \} | \quad  \forall t \\
    a_t + d_t = | \{ n : z_{tn} > 0 \}| \quad \forall t \\
  \end{cases}
  \end{aligned}
  \label{eqn:z}
\end{equation}
Association hypotheses that do not satisfy these constraints have zero
probability. We note that the space of possible associations is
exponential in time $T$ and factorial in the number of observations $N_t$ at
each time $t$ \cite{pasiliao2010local}. Reasoning over this large space is \emph{the} fundamental challenge
in data association. Doing so while satisfying these constraints makes inference
difficult, as we discuss in Section~\ref{sec:inference}.

\subsection{Dynamics, Observations $p(x \mid z)\ p(y \mid x, z)$}
Denote the trajectory of object $k>0$ at time $t$ by the latent random variable
$x_{tk} \in \mathbb{R}^{D_x}$ for $D_x$ the dimension of the latent states.
Denote all trajectories as $x = \{ x_1, \ldots, x_T \}$ where $x_t = \{ x_{tk}
\}_{k=1}^{K(z)}$ for $K(z)$ the number of objects in association hypothesis $z$.
Every object has a length-$T$ latent trajectory, but is represented by $x_{tk} =
\emptyset$ for any time before its arrival or after its departure. Let $t' =
t-1$ and define the dynamics model for objects $1, \ldots, K(z)$ as:
\begin{equation}
  \begin{aligned}
  p(x \mid z) &= \prod_{t=1}^T p(x_t \mid x_{1:t-1}, z)
  = \prod_{t=1}^T \prod_{k=1}^{K(z)}
  p(x_{tk} \mid x_{t' k})
  \end{aligned}
  \label{eqn:x}
\end{equation}
Define the observation model as,
\begin{equation}
    p(y \mid x, z) = \prod_{t=1}^T p(y_t \mid z_t, x_t)
    = \prod_{t=1}^T \prod_{n=1}^{N_t}
      p(y_{tn} \mid z_{tn}, x_t)
    \label{eqn:y}
\end{equation}
We now specialize Eqns.~\ref{eqn:x}, \ref{eqn:y} to a linear Gaussian system, as
is common in tracking. For the dynamics,
\begin{equation}
\begin{aligned}
  p(x_{tk} \mid x_{t' k}) =
  \begin{cases}
    \text{N}\left( x_{tk} \mid
      F x_{t' k},
      Q
    \right) 
    &\text{ if  } x_{t' k} \neq \emptyset 
    \\
    \text{N}\left( x_{tk} \mid
      \mu_0,
      \Sigma_0
    \right) 
    &\text{ o.w. }
  \end{cases}
  \label{eqn:xtk}
\end{aligned}
\end{equation}
The first line is a linear Gaussian system with system model $F$
and noise covariance $Q$. The second line specifies a shared prior on
trajectories with prior parameters $\mu_0, \Sigma_0$ that are typically set to
be broad over the observation space. JPT marginalizes over missing detections
(i.e., times when an object has already arrived but has no association). This
can be computed in closed-form for linear Gaussian dynamics.

For the observation model,
\begin{equation}
  p(y_{tn} \mid z_{tn}=k, x_t) = \begin{cases}
    \text{N}\left( y_{tn} \mid
      H x_{tk},
      R
    \right)
    &\text{ if } k > 0 \\
    \text{N}\left( y_{tn} \mid
      \mu_{\text{FP}},
      \Sigma_{\text{FP}}
    \right)
    &\text{ o.w. }
  \end{cases}
  \label{eqn:ytk}
\end{equation}
The first line is a linear Gaussian system with observation projection $H$ and
observation noise covariance $R$. The second line specifies the model for
clutter detections with prior parameters $\mu_{\text{FP}}, \Sigma_{\text{FP}}$,
which are also typically set to be broad.

\subsection{Joint Distribution}
Finally, the joint posterior over trajectories $x$, associations $z$ and
counts $M$ given observations $y$ is,
\begin{equation}
  p(x, z, M \mid y) = \frac{1}{Z} p(M) \, p(z \mid M)\, p(x \mid z)\, p(y \mid x, z)
  \label{eqn:posterior}
\end{equation}
where each term on the RHS is respectively given by Equations~\ref{eqn:M},
\ref{eqn:z}, \ref{eqn:x}, \ref{eqn:y} and $Z$ is an intractable normalization
constant (owing to the exponential and factorial number of terms). We show how
to draw samples from this non-trivial posterior using Metropolis-Hastings
proposals in Section~\ref{sec:inference}.

\section{JPT Inference}
\label{sec:inference}
Sampling from the posterior in Eqn.~\ref{eqn:posterior} is
complicated by the constraints in Eqn.~\ref{eqn:z} and the exponential in
$T$ and factorial in $N_t$ scaling of possible association hypotheses. With
no analytic form and computationally infeasible enumeration of all hypotheses,
we turn to Metropolis-Hastings \cite{hastings1970monte}.

The Metropolis-Hastings (MH) algorithm enables sampling from intractable
distributions by constructing a Markov chain whose unique
stationary distribution is the desired distribution. Samples from this chain
converge in distribution to the desired distribution, regardless of starting
state. MH constructs transition distributions $q^*$ that maintain detailed balance,
\begin{equation}
  \frac{
    p(x', z', M' \mid y)\
  }{
    p(x, z, M \mid y)
  }
  =
  \frac{
    q^*(x', z', M' \mid x, z, M, y)
  }{
    q^*(x, z, M \mid x', z', M', y)
  }
\end{equation}
resulting in a chain where Equation~\ref{eqn:posterior} is a stationary distribution.
MH accepts a proposed sample $(x', z', M')$ from an arbitrary 
proposal distribution $q$ with probability $\min(1, R)$,
\begin{equation}
    R = \frac{p(x', z', M' \mid y)}{p(x, z, M \mid y)}
    \ 
    \frac{q(x, z, M \mid x', z', M', y)}{q(x', z', M' \mid x, z, M, y)}.
    \label{eqn:MH}
\end{equation}
where the normalizers cancel. In Section~\ref{sec:proposals}, we design
proposals that rapidly explore high-probability regions in the JPT posterior,
hopping between different modes as demonstrated in Figure~\ref{fig:teaser} and
later quantified in Section~\ref{sec:experiments}.  We then describe closed-form
Gibbs sampling of joint trajectories conditioned on associations in
Section~\ref{sec:ffbs}.
\subsection{Proposals}
\label{sec:proposals}
\begin{figure}[]
  \includegraphics[width=0.95\textwidth]{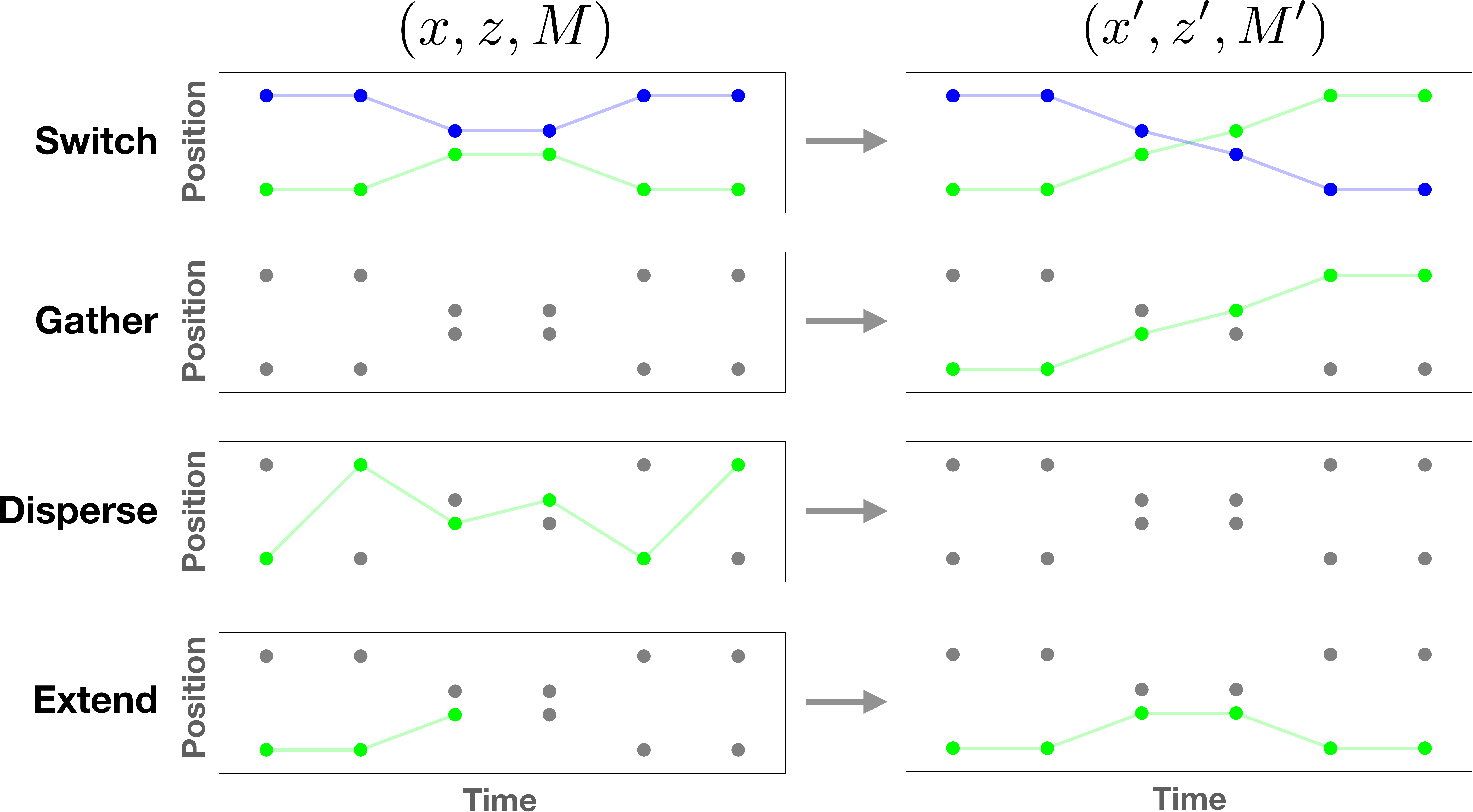}
  \caption{Examples of each JPT proposal. Left column is the
  input state $(x, z, M)$ and right column is the output state $(x', z', M')$.
  Black points are observations $y$, encircled in the color of their association
  (green or blue for objects; grey for clutter). Example trajectories $x$ are
  visualized as colored lines.  Switch proposals can reason over many objects,
  but are shown here for two.
  }
  \label{fig:proposals}
\end{figure}
We design Metropolis-Hastings proposals that make large moves in the latent
space (including mode hopping) by reasoning over permutations of the latent
state over time. JPT proposals are data-dependent--they make use of the
observations $y$ and current state $(x, z, M)$ in proposing next state $(x', z',
M')$. Data-dependent proposals are complicated but in our case avoid random
exploration and the use of gating heuristics while retaining tractability.
Broadly, JPT proposals reason over assignment and trajectory permutations
between existing objects (Switch \ref{sec:switch}), between a new object and
clutter (Gather \ref{sec:gather}, Disperse \ref{sec:disperse}) or between existing
objects and clutter (Extend, \ref{sec:extend}). Of these, the Switch proposal is a
novel generalization of \cite{neal2004inferring} and contributes most to JPT's
exploration of posterior modes; we thus focus on it more than the other
proposals. Pictorial examples for each proposal transitioning from state $(x, z,
M)$ to state $(x', z', M')$ are shown in Figure~\ref{fig:proposals}.
%
%
\subsubsection{Switch Proposal}
\label{sec:switch}
\begin{algorithm}[h]
  \DontPrintSemicolon
  \SetKwInOut{Input}{Input}
  \SetKwInOut{Output}{Output}
  \Input{$\ x, z, M, y$}
  \Output{$\ x', z', M'$}
  Let $x' = x,\ z' = z$ \\
  Sample object set $\mathcal{K} \subset \{ 1, \ldots, K(z) \}$ s.t.
  $|\mathcal{K}| \geq 2$ \label{alg:switch:line:K}\\
  Define switch times $\tau = \{ t : z_{tn} = k \text{ for any } k \in \mathcal{K} \}$ \label{alg:switch:line:tau} \\
  Set permutations $\sigma_t$ as the identity permutation on $(1, \ldots, K(z))$
    for any $t \not \in \tau$  \label{alg:switch:line:identity_perm} \\
  \For{$t \in \tau$ in order}
  {
    Sample valid permutation
    $p(\sigma_t \mid \sigma_{1:t-1}) \propto
      p\left(\sigma_t(x_t) \mid \sigma_{1:t-1}(x_{1:t-1})\right)\ p\left(y_t \mid
      \sigma_t(x_t)\right)
    $ 
    \label{alg:switch:line:sigma_sample}
    \\
    Let $x_t' = \sigma_t(x_t),\ z_t' = \sigma_t(z_t)$
  }
  Compute counts $M'$ from $z'$ \label{alg:switch:line:counts} \\
  \lIf{$M' = M$ \text{ or } $\text{rand}(0,1) < \min(1, R_{\text{switch}})$} {
    return $x', z', M'$
  }\label{alg:switch:line:mh}
  \lElse{
    return $x, z, M$
  } 
  \caption{Switch Proposal}
  \label{alg:switch}
\end{algorithm}
Switch proposals consider possible trajectory and associations permutations
between existing objects, and are sampled according to JPT's dynamics and
observation models (Eqns.~\ref{eqn:x}, \ref{eqn:y}). They cause rapid
exploration of different posterior modes such as the ones shown in
Figure~\ref{fig:teaser}.  Strikingly, the Switch proposal is in many cases
automatically accepted ($R_{\text{switch}}=1$).

Following Algorithm~\ref{alg:switch}, the Switch proposal samples uniformly at 
random a subset $\mathcal{K}$ of existing objects $\{1, \ldots, K(z)\}$ such
that $2 \leq |\mathcal{K}| \leq \bar{\mathcal{K}}$
(Line~\ref{alg:switch:line:K}) for $\bar{\mathcal{K}}$ a maximum size, discussed below.

Let $\sigma_t$ be a \emph{valid permutation} on objects $\{1, \ldots, K(z)\}$ at
time $t$. Valid permutations do not permute objects outside of $\mathcal{K}$:
for all $k \not \in \mathcal{K}, \sigma_t(k) = k$. With slight abuse of
notation, let $\sigma_t(x_t)$ and $\sigma_t(z_t)$ respectively represent the
trajectory values and associations permuted according to $\sigma_t$. So for time
$t$, the trajectory value $x_{tk}$ (possibly an uninstantiated value) and
association (possibly none) of object $k$  become the trajectory value and
association of object $\sigma_t(k)$. Define $\sigma_{1:t}(x_{1:t})$ over times
$1, \ldots, t$ as $x_{1:t}'$ where $x_t' = \sigma_t(x_t)$.

The Switch proposal only considers permutations at times when at least one 
object $k \in \mathcal{K}$ has been observed. Let $\tau$ be all such times 
(Line~\ref{alg:switch:line:tau}). For any time $t \not 
\in \tau$, set $\sigma_t$ as the identity permutation, $\sigma_t(k) = k$
(Line~\ref{alg:switch:line:identity_perm}). 

For increasing time $t \in \tau$, iteratively sample
permutation $\sigma_t$ conditioned on the previously-sampled permutations
$\sigma_{1:t-1}$ with probability proportional to the product of the observation
and dynamics models (Equations~\ref{eqn:x}, \ref{eqn:y}) evaluated with the 
appropriate swaps in trajectory and association values imposed by permutations 
$\sigma_{1:t}$ (Line~\ref{alg:switch:line:sigma_sample}).
There are $|\mathcal{K}|!$ possible values for $\sigma_t$ at each time $t$, but
we find $\bar{\mathcal{K}} = 7$ balances efficient computation and posterior
exploration.

After sampling $\sigma_t$ for all $t \in \tau$, we compute new counts $M'$ from
the permuted associations $z'$ (Line~\ref{alg:switch:line:counts}) and the
Hastings ratio (Line~\ref{alg:switch:line:mh}) between $(x', z', M')$ and $(x,
z, M)$, noting that Switch proposals are their own reverse move. In
Appendix~\ref{sec:switch_ratio}, we show that:
\begin{equation}
  R_\text{switch} = 
  \frac{
    \prod_{t=1}^T
    p(M_t' \mid M_{t-1}')
  }{
    \prod_{t=1}^T
    p(M_t \mid M_{t-1})
  }.
  \label{eqn:switch:mh}
\end{equation}
The Switch proposal is always accepted 
($R_\text{switch}=1$) whenever $M' = M$. This occurs in several
situations: when the number of objects are known in advance, when objects are
assumed never to depart, when there are no missing observations and when all $k
\in \mathcal{K}$ are observed at $\max \tau$. In many scientific and sports
analytics applications, it is common for subjects to never depart. When these
conditions don't hold, the event counts and Hastings ratio are efficiently
evaluated (linear in time $T$ and parallelizable) by only considering terms
where the counts $M', M$ differ. Switch proposals have complexity
$\mathcal{O}(|\mathcal{K}|!\ T)$ where the factorial dependence on
$|\mathcal{K}|$ comes from Line~\ref{alg:switch:line:sigma_sample} and the
linear dependence on $T$ comes from its enclosing for loop. In practice, we
limit the subset size $|\mathcal{K}| \leq \bar{\mathcal{K}}$.

In Appendix~\ref{sec:switch_ehmm}, we show that Switch proposals generalize
the Extended HMM proposals of \cite{neal2004inferring} by proposing a
discretization that depends on the current latent state (in their nomenclature,
JPT "pool states" are permutations of $x, z$). In their work, sampled
discretizations (or pool states) cannot depend on the current latent state, else
detailed balance is lost.
\subsubsection{Gather Proposal}
\label{sec:gather}
\begin{algorithm}[]
  \DontPrintSemicolon
  \SetKwInOut{Input}{Input}
  \SetKwInOut{Output}{Output}
  \Input{$\ x, z, M, y$}
  \Output{$\ x', z', M'$}
  Let $x' = x,\ z' = z$ \\
  Let $k = 1 + K(z)$ \label{alg:gather:k} \\
  Define gather times
    $\tau_0 = \{ t : z_{tn} = 0 \text{ for any } 1 \leq t \leq T \}$
    \label{alg:gather:tau0} \\
  \For{$t = \min \tau_0, \ldots, \max \tau_0$}{
    \lIf{$\text{rand}(0,1) < \delta$} {
      continue
    } \label{alg:gather:delta}
    Sample
    $ \ \ 
      p(z_{tn}' = k) \propto p(y_{tn} | x_{t' k}, z_{tn}'=k)\ \mathbb{I}(z_{tn} = 0)
    $
    \label{alg:gather:z} \\
    Sample 
    $ \ \ 
      p(x_{tk}' \mid x_{t' k}', z_{tn}'=k) \propto {p(x_{tk}' \mid x_{t' k})\ p(y_{tn} \mid x_t', z_{tn}'=k)}
    $
    \label{alg:gather:x}
  }
  Compute counts $M'$ from $z'$ \\
  \lIf{$\text{rand}(0,1) < \min(1, R_{\text{gather}})$}{
    return $x', z', M'$
    \label{alg:gather:mh}
  }
  \lElse{
    return $x, z, M$
  } 
  \caption{Gather Proposal}
  \label{alg:gather}
\end{algorithm}
Following Algorithm~\ref{alg:gather}, the Gather proposal considers the
formation of a new object $k = 1+K(z)$ (Line~\ref{alg:gather:k}) from the set of
clutter-associated observations $\{y_{tn} : z_{tn} = 0\}$. Its reverse move
is Disperse (\ref{sec:disperse}). Let $\tau_0$ be the set of times $t$ with at 
least one clutter association (Line~\ref{alg:gather:tau0}). 
For increasing $t \in \tau_0$, assignments to object $k$ are
iteratively sampled either among observations that are currently associated to 
clutter or, with probability $\delta=0.01$, no clutter
association to allow for missing observations.  
In the former, association $z_{tn}'=k$ is sampled among all clutter observations
with probability proportional to Line \ref{alg:gather:z} where $t' = t-1$ and
marginalization occurs between states with missing associations.

Conditioned on the sampled assignment $z_{tn}'=k$, a trajectory value $x_{tk}'$
is sampled (Line~\ref{alg:gather:x}); this is analytic in the linear Gaussian
case.  Sampling the association then the trajectory allows unambiguous
evaluation of the reverse move in the Hastings ratio (Line~\ref{alg:gather:mh})
as derived in Appendix~\ref{sec:gather_ratio}. Gather always requires an
accept/reject step and has complexity $\mathcal{O}(T\ \prod_{t=1}^T N_t)$, where
the linear complexity in $N_t$ is a consequence of
Lines~\ref{alg:gather:z}--\ref{alg:gather:x} and linear complexity in $T$
comes from the enclosing for loop.

\subsubsection{Disperse Proposal}
\label{sec:disperse}

\begin{algorithm}[]
  \DontPrintSemicolon
  \SetKwInOut{Input}{Input}
  \SetKwInOut{Output}{Output}
  \Input{$\ x, z, M, y$}
  \Output{$\ x', z', M'$}
  Let $z' = z$ \\
  Sample $k \in \{1, \ldots, K(z)\}$ \label{alg:disperse:sample} \\
  Set $z_{tn}' = 0$ for all $t,n$ such that $z_{tn} = k$ \label{alg:disperse:z} \\
  Let $x' = x \setminus \{ x_{tk} \}_{t=1}^T$ \label{alg:disperse:x} \\
  Compute counts $M'$ from $z'$ \\
  \lIf{$\text{rand}(0,1) < \min(1, R_{\text{disperse}})$} {
    return $x', z', M'$
  }
  \lElse{
    return $x, z, M$
  } \label{alg:disperse:mh}
  \caption{Disperse Proposal}
  \label{alg:disperse}
\end{algorithm}
Following Algorithm~\ref{alg:disperse}, the Disperse proposal simply chooses an
existing object at random (Line~\ref{alg:disperse:sample}), removes all its
associations by setting them to clutter (Line~\ref{alg:disperse:z}) and deletes
the trajectory values for that object (Line~\ref{alg:disperse:x}). It is the
reverse move for the Gather proposal. Hence, $R_\text{disperse} =
R_\text{gather}^{-1}$ where $R_\text{gather}$ is defined in Equation
\ref{eqn:gather:mh:line2}.  As in the case for the Gather proposal, an
accept/reject step is required. Disperse has constant complexity.

\subsubsection{Extend Proposal}
\label{sec:extend}

\begin{algorithm}[]
  \DontPrintSemicolon
  \SetKwInOut{Input}{Input}
  \SetKwInOut{Output}{Output}
  \Input{$\ x, z, M, y$}
  \Output{$\ x', z', M'$}
  Let $x' = x,\ z' = z$ \\
  Sample $k \in \{1, \ldots, K(z) \}$ \label{alg:extend:sample} \\
  Define extend times
    $\tau_k = \{ t : z_{tn} \in \{ 0, k \} \text{ for any } 1 \leq t \leq T \}$ \label{alg:gather:tauk} \\
  \For{$t = \min \tau_k, \ldots, \max \tau_k$}{
    \lIf{$\text{rand}(0,1) < \delta$} {
      continue
    } \label{alg:extend:delta}
    Sample
    $ \ \ 
      p(z_{tn}' = k) \propto p(y_{tn} | x_{t' k}, z_{tn}'=k)\ \mathbb{I}(z_{tn}
      \in \{0, k\})
    $ 
    \label{alg:extend:z} \\
    Sample 
    $ \ \ 
      p(x_{tk}' \mid x_{t' k}', z_{tn}'=k) \propto p(x_{tk}' \mid x_{t' k})\ p(y_{tn} \mid x_t', z_{tn}'=k)
    $
    \label{alg:extend:x}
  }
  Compute counts $M'$ from $z'$ \\
  \lIf{$\text{rand}(0,1) < \min(1, R_{\text{extend}})$} {
    return $x', z', M'$
  }
  \lElse{
    return $x, z, M$
  } \label{alg:extend:mh}
  \caption{Extend Proposal}
  \label{alg:extend}
\end{algorithm}
The Extend proposal is similar to the Gather proposal but rather than consider
permutations between clutter associations and a new object, it considers
permutations between clutter associations and an existing object. Effectively,
this allows an existing object to resample associations.

Following Algorithm~\ref{alg:extend}, randomly sample object $k$ from existing
objects (Line~\ref{alg:extend:sample}) and iterate over all times $t \in \tau_k$
with an association to clutter $z_{tn} = 0$ or to the current object $z_{tn} = k$
(Line~\ref{alg:gather:tauk}). As in the Gather proposal, skip a resampling of
assignments at time $t \in \tau_k$ with probability $\delta$. Otherwise, sample
an association then a trajectory value
(Lines~\ref{alg:extend:z}-\ref{alg:extend:x}) with definitions as in Gather,
except that it is possible for $z_{tn}' = z_{tn}$ for some times $t$ (it
resamples the same assignment it already had).

By automatically rejecting any Extend proposal that leaves a object with fewer
than two observations, we can ensure that Extend proposals are always their own
reverse move. As in Gather, the observation and dynamics terms cancel
in the posterior ratio for all objects other than $k$, but an accept/reject step
must still be computed, and is of similar form to the Gather proposal. Like
Gather, Extend has complexity $\mathcal{O}(T\ \prod_{t=1}^T N_t)$ where linear
complexity in $N_t$ comes from Lines~\ref{alg:extend:z}--\ref{alg:extend:x} and
linear complexity in $T$ comes from the enclosing for loop.

\subsection{Forward-Filtering, Backward Sampling}
\label{sec:ffbs}
Joint sampling of trajectories from the full conditional,
\begin{equation}
  p(x \mid z, M, y) = p(x \mid z, y)
\end{equation}
constitutes a fifth MH proposal in the form of a Gibbs sampler where $M$ is
dropped due to independence. As discussed, jointly sampling $x \mid z, y$
differs from typical filter- and smoothing-based approaches. If there are no
states with missing associations, then the full conditional on trajectories can
be sampled as,
\begin{align}
  & p(x \mid z, y)
  = \prod_{k=1}^{K} \prod_{t=1}^T
    \frac{
      p(x_{tk} \mid y_{1:t}^k) \ p( x_{(t+1)k} \mid x_{tk} )
    }{
      p(x_{(t+1)k} \mid y_{1:t}^k)
    }
    \label{eqn:ffbs}
\end{align}
where $p(x_{tk} \mid y_{1:t}^k)$ is the filter distribution of $x_{tk}$ and
$y_{1:t}^k = \{ y_{t' n} : z_{t' n} = k \text{ and } t' \leq t \}$. Sampling
from this posterior is similar to smoothing \cite{rauch1965maximum}, except
that the backwards pass draws samples. Inference can be done in parallel over objects
and, in the linear Gaussian case, is in closed form with complexity linear in
$T$. For other (possibly non-linear) dynamics or observation models, any
procedure that leaves the joint distribution invariant may be used. Described in
Appendix~\ref{sec:ffbs_missing}, we marginalize over latent states at times when the object has no
association.

%
\section{Experiments}
\label{sec:experiments}
Our experiments demonstrate that JPT provides a superior representation of
posterior uncertainty as compared to batch MCMCDA (\ref{sec:uq}) by virtue of
more thoroughly and efficiently exploring the configuration space. Following, we
show that JPT outperforms MCMCDA and a recent, optimization-based tracker on
large datasets (\ref{sec:mot_metrics}). Lastly, we show that JPT facilitates
targeted queries to an oracle (e.g., a noisy human annotator) yielding
significant improvement in trajectory quality with fewer iterations
(\ref{sec:uq_reduction}).

In all experiments, JPT and MCMCDA are run for $5$ replicates (Markov chains),
each drawing $2000$ samples and discarding half as burn-in. All
approaches use the same linear Gaussian dynamics.

\subsection{Representation of Posterior Uncertainty}
\label{sec:uq}
\begin{figure}
  \centering
  \includegraphics[width=0.9\textwidth]{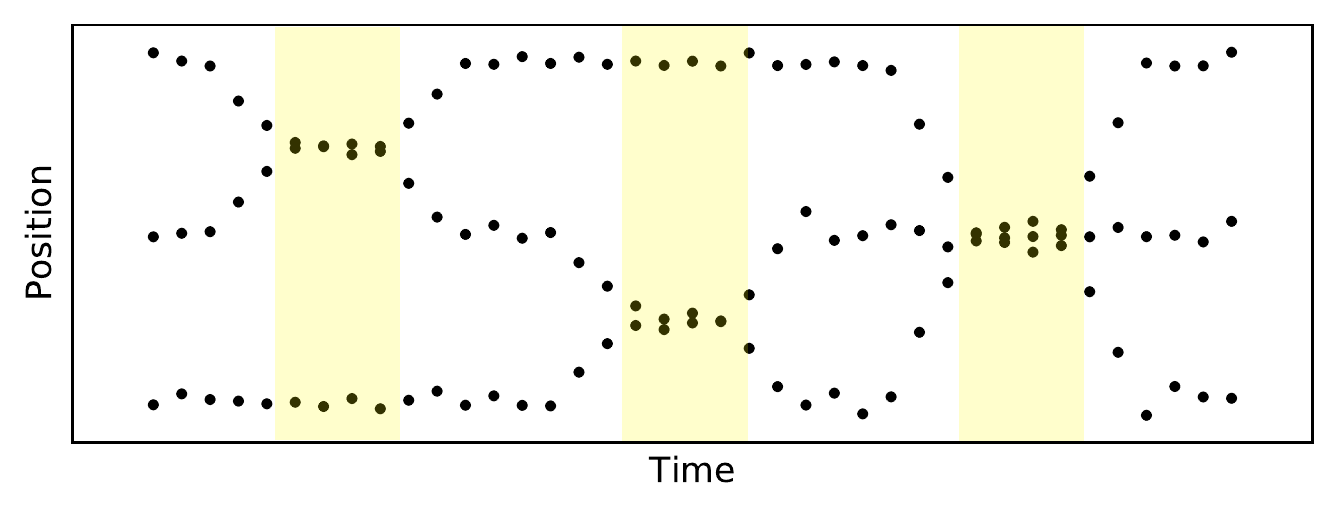}
  \includegraphics[width=0.9\textwidth]{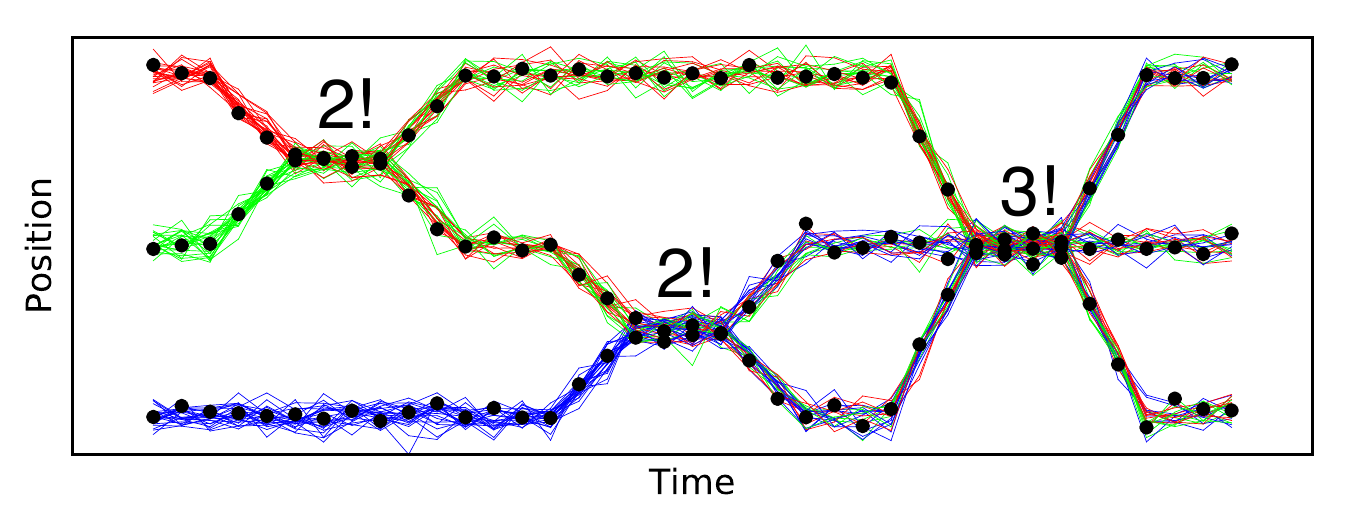}
  \caption{The K33 dataset. Observations over time (\textbf{Top}) with yellow shading for
  ambiguous regions and a joint trajectory sample from the $24 = 2!\ 2!\ 3!$
  posterior modes (\textbf{Bottom}), each reflecting a possible outcome.}
  \label{fig:k33_dataset}
\end{figure}
Consider the K33 dataset shown in Figure~\ref{fig:k33_dataset} (Top). Three objects
begin well-separated but become ambiguous after each of three confusion events
(yellow shading). Observe that for $k$ ambiguously proximate objects, there will
be $k!$ possible outcomes. Figure~\ref{fig:k33_dataset} (Bottom) shows the $24 =
2!\ 2!\ 3!$ modes that a multi-object tracker would ideally explore in this
dataset.

\begin{figure}
  \centering
  \includegraphics[width=0.9\textwidth]{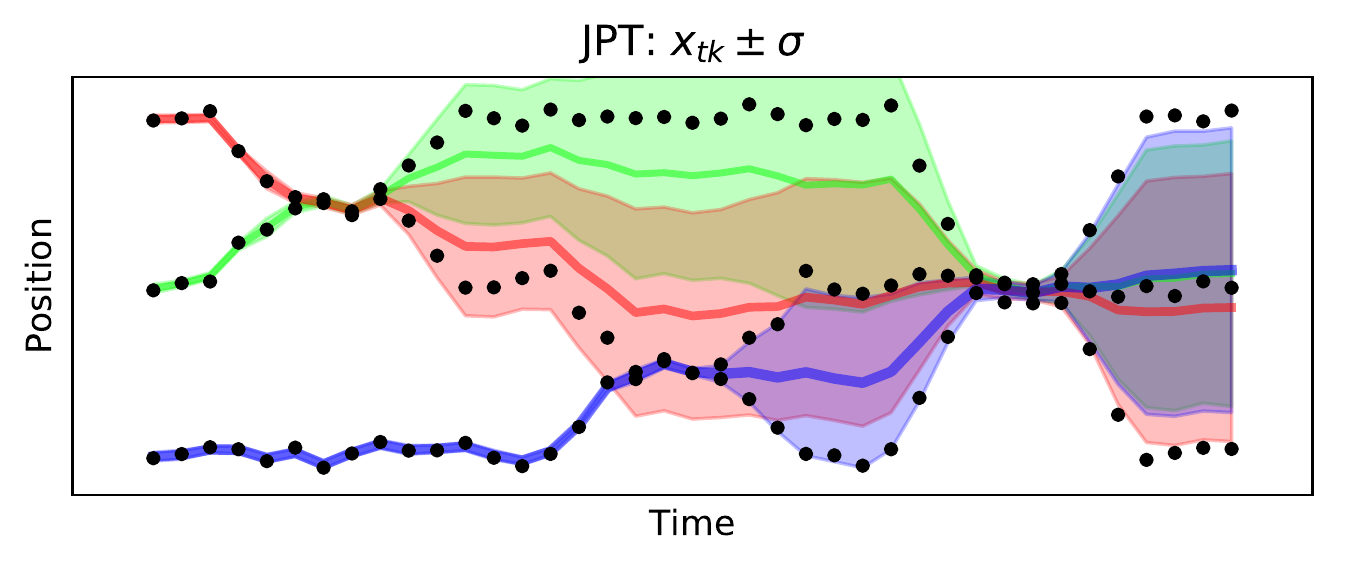}
  \includegraphics[width=0.9\textwidth]{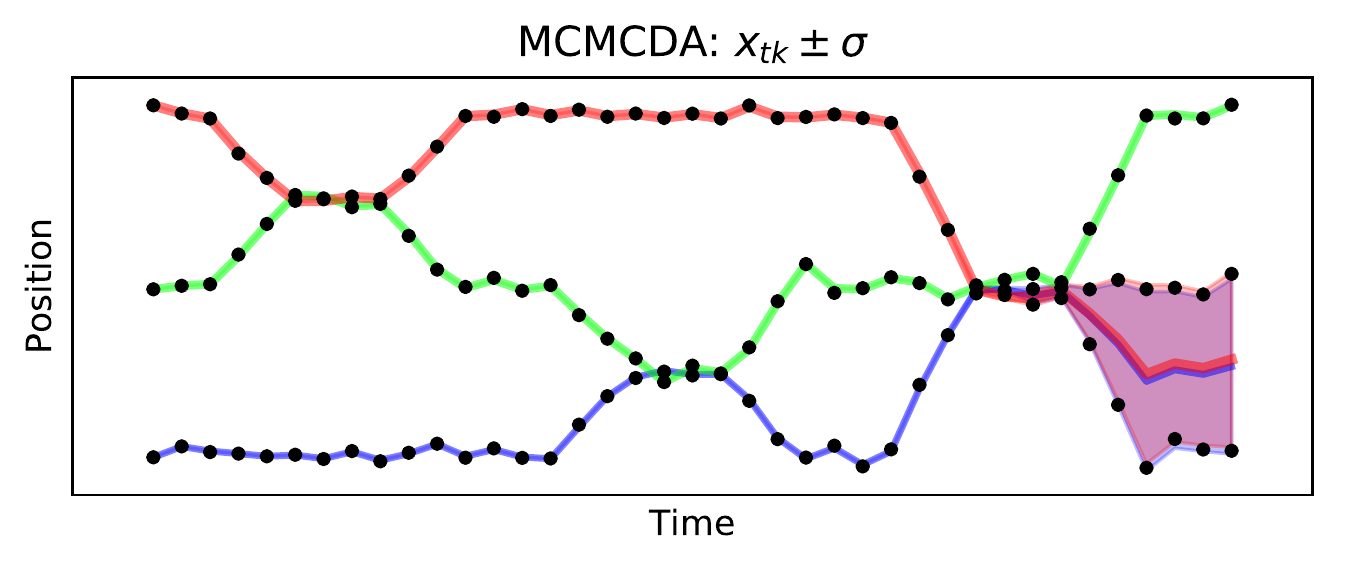}
  \caption{Posterior trajectory values $\pm$ one SD for JPT
  (\textbf{Top}) and MCMCDA (\textbf{Bottom}). JPT correctly captures the uncertainty from each
  ambiguous regions while MCMCDA fails to represent most ambiguities.}
  \label{fig:k33_xtk}
\end{figure}
We investigate whether JPT and MCMCDA effectively explore the $24$ modes of the
K33 dataset by observing posterior trajectory variance for one Markov chain in
Figure~\ref{fig:k33_xtk}. JPT captures the uncertainty arising in each ambiguous
region; for example, red and green cross in some posterior samples but not in
others. In contrast, MCMCDA is \emph{overconfident}: it fails to represent any
uncertainty in either of the first two ambiguous regions, and is only partially
successful in the final region. Not only is MCMCDA
overconfident, it is wrong as will be shown by tracking metrics in
(\ref{sec:mot_metrics}).


\begin{figure}
  \centering
  \includegraphics[width=0.9\textwidth]{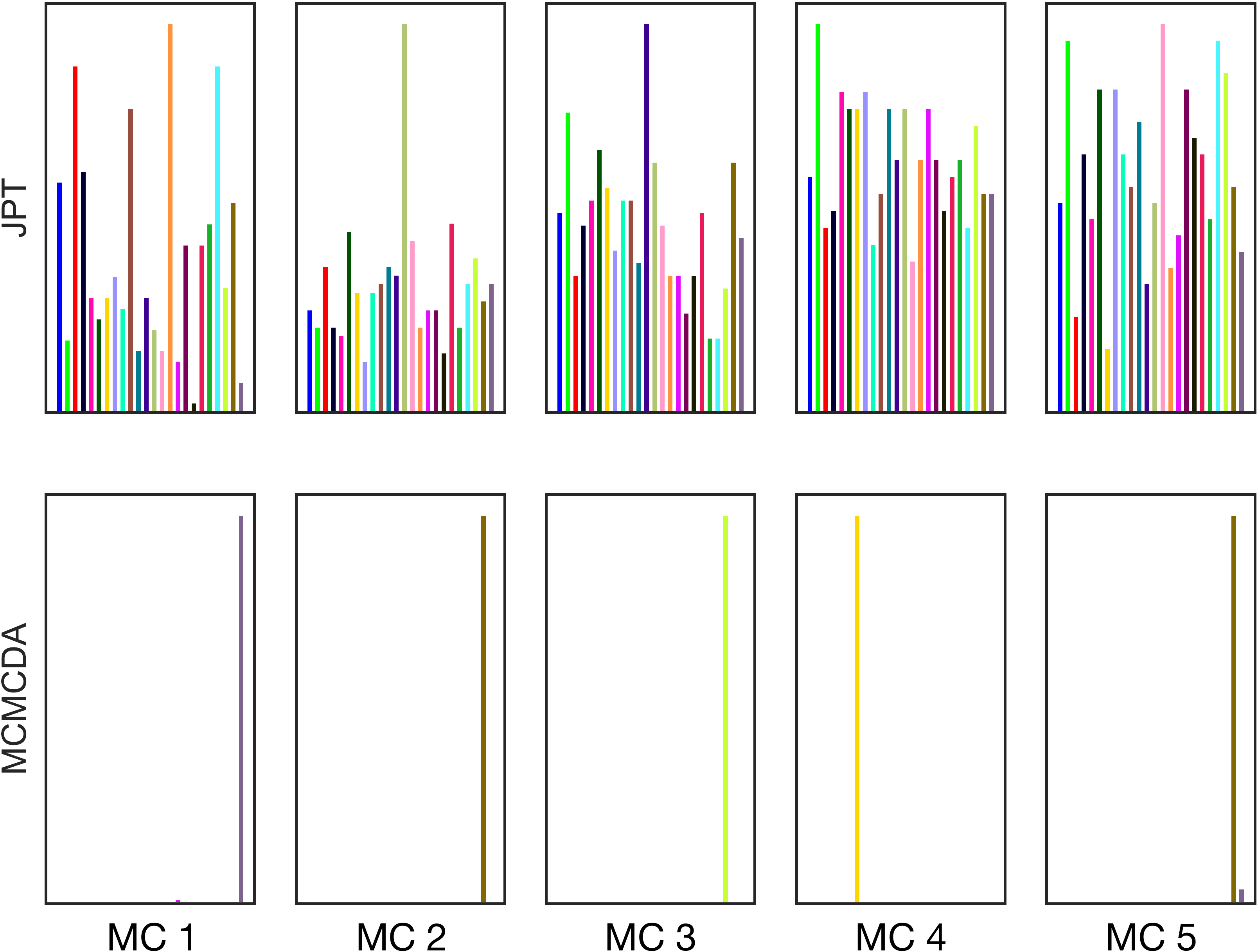}
  \caption{
    Histograms of the modes captured by JPT (\textbf{Top}) and MCMCDA
    (\textbf{Bottom}) in $5$ Markov chains (MC). JPT explores all modes in each
    chain while MCMCDA gets stuck in one or two.}
\label{fig:k33_hist}
\end{figure}
\begin{figure}
  \centering
  \includegraphics[width=0.7\textwidth]{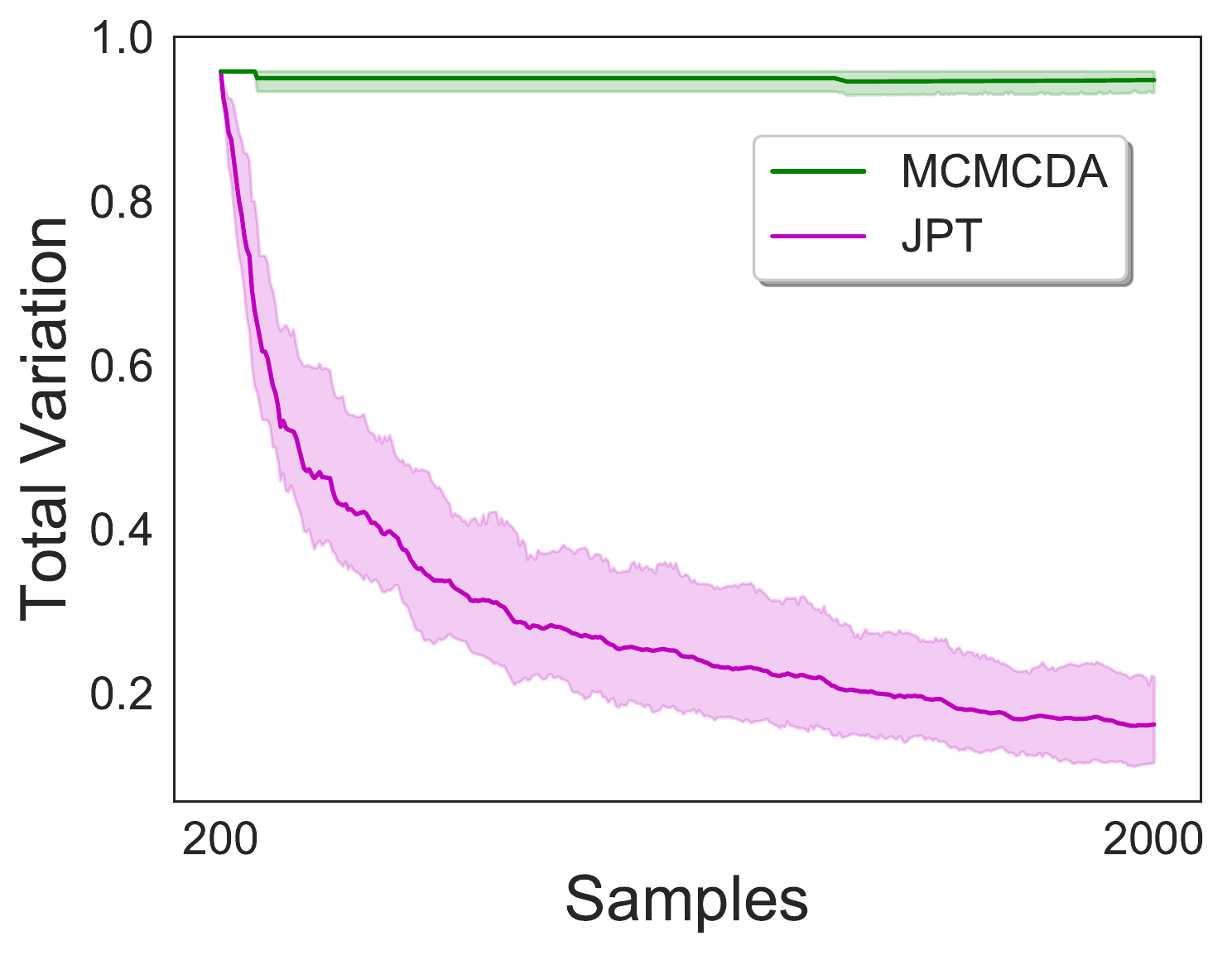}
  \caption{
  Total variation distance between the true distribution of modes on K33, and
  the histograms of matched modes for JPT and MCMCDA samples.
}
\label{fig:k33_total_variation}
\end{figure}
We quantify how well uncertainty is captured by matching each posterior sample
from JPT and MCMCDA to the nearest of the $24$ likely outcomes. Details of this
matching procedure are in Appendix~\ref{sec:assoc_distance}.  Figure~\ref{fig:k33_hist} shows
histograms of modes matched by JPT and MCMCDA in different Markov chains. JPT
represents every outcome within each Markov chain while MCMCDA captures at most
$2$ but usually $1$ outcome in a single Markov chain. Noting that the ideal
distribution over the $24$ matched modes would be uniform, we compare total
variation (L1) distance between that and the empirical distributions of
matched modes for JPT and MCMCDA (Figure~\ref{fig:k33_total_variation}), plotted
as a function of sample count. Observe that JPT's total variation is low but
nonzero, implying imperfect mode exploration. This can be seen in
Figure~\ref{fig:k33_xtk} (Top) where the red and green means are not perfectly
balanced between the upper and middle paths after the first ambiguous region.
Although not uniform in its mode exploration, JPT captures each outcome and is
dramatically closer to the ideal uniform distribution than MCMCDA.

\subsection{Performance on Real and Synthetic Data}
\label{sec:mot_metrics}
\begin{figure}[]
  \centering
  \includegraphics[width=\textwidth]{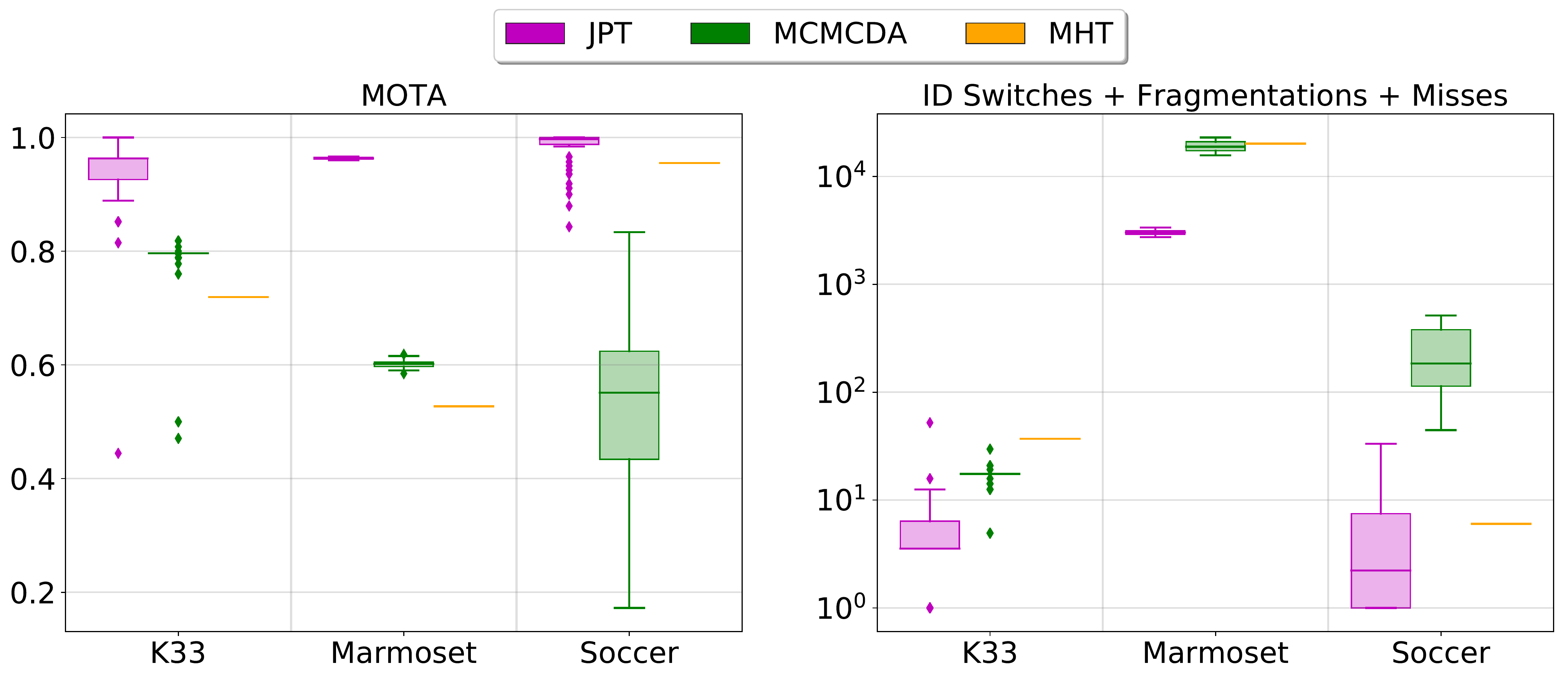}
  \caption{CLEAR MOT metrics for JPT, MCMCDA and MHT on datasets K33, Marmoset
  and Soccer. (\textbf{Left}), higher is better; (\textbf{Right}), lower is
  better.}
  \label{fig:mot_metrics}
\end{figure}
We compare tracking performance of JPT, MCMCDA and a modern optimization-based
tracker \cite{kim2015multiple} (MHT) on three datasets: the K33 dataset (39
timesteps, 3 objects, 117 observations), a scientific dataset Marmoset (15k
timesteps, 2 objects, 25k observations), and the sports dataset Soccer (1.5k
timesteps, 22 objects, 12k observations). Metrics are computed over $200$
evenly-spaced samples after burn-in for JPT and MCMCDA. Being deterministic, MHT
only provides a point estimate. Details of each dataset are in the
Appendix~\ref{sec:datasets}.

Figure~\ref{fig:mot_metrics} shows performance as evaluated by standard CLEAR
MOT \cite{bernardin2008evaluating} metrics that account for identity switches
(objects get confused), fragmentations (two inferred objects explain one actual
object) and misses (an observation isn't correctly associated to an object).

Multi-object tracking accuracy (MOTA) is a summary statistic accounting for
these events. JPT outperforms MCMCDA and MHT on all datasets and metrics with
notably fewer identity switches, fragmentations and misses. 
Because MCMCDA uses gating heuristics, the reported performance is taken 
as the best-scoring MOTA from a grid search over parameter values for each heuristic.
Details of the grid search are in Appendix~\ref{sec:mcmcda_grid}.

\subsection{Uncertainty Reduction}
\label{sec:uq_reduction}

\begin{figure}
  \centering
  \includegraphics[width=0.45\textwidth]{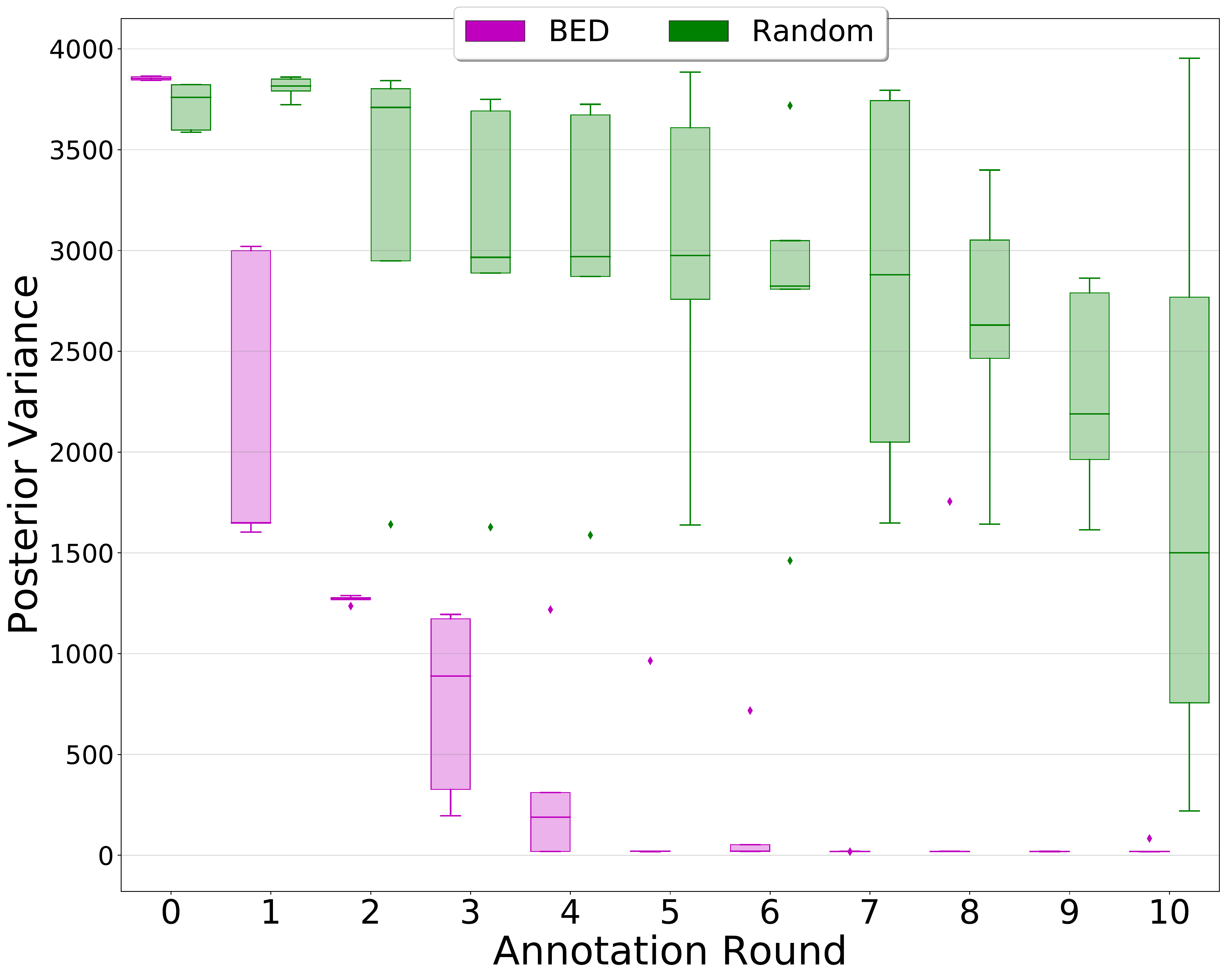}
  \hspace{1cm}
  \includegraphics[width=0.45\textwidth]{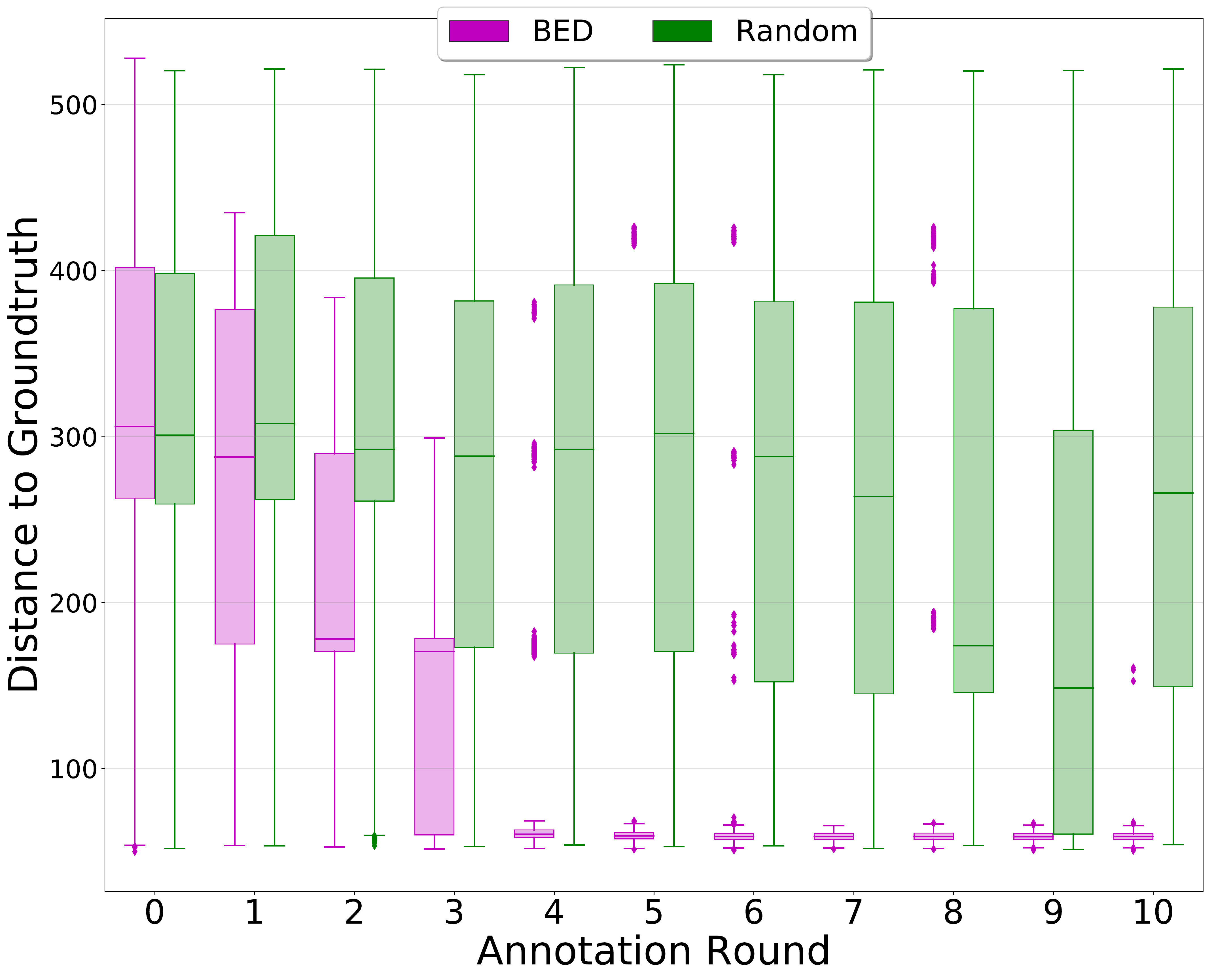}
  \caption{Successive rounds of automatically-scheduled annotations reduce
  posterior uncertainty (left) and improve trajectory quality (right) when
  planning with JPT's uncertainty representation (magenta) as compared to
  planning with no model of uncertainty (green).}
  \label{fig:k33_uq_reduction_quality_improvement}
\end{figure}

%
We show JPT's accurate representation of posterior uncertainty facilitates 
follow-on analysis. Specifically, it provides a pathway to obtain high quality 
trajectories despite significant ambiguity in the original observation set. 
We augment the original dataset with
disambiguations or \emph{annotations} that have the potential to resolve modes in 
the posterior. Uncertainty representation allows us to identify 
\emph{informative} annotations, thereby requiring few of them to arrive at
quality trajectories.

Let there be $L$ annotations $a = \{ a_l \}_{l=1}^L$ where each indicates
whether two observations $y_{t_1 n_1}, y_{t_2 n_2}$ belong to the same or
different objects: $a_l(y_{t_1 n_1}, y_{t_2 n_2}) =1 \text{ or } 0$.  Assume
each annotation is correct with probability $p_a = 0.99$. Intuitively,
\emph{informative} annotations involve observations that flank an ambiguous
region since the annotated value supports only a subset of modes in the
posterior, thereby reducing uncertainty.  We use sequential Bayesian
experimental design (BED) for automated selection of informative observation
pairs for $L$ annotation rounds. 

Sequential BED iteratively chooses observation pairs for annotation that yields the
greatest information about the latent trajectories, as measured by mutual 
information (MI) \cite{bernardo1979mi}. MI quantifies the expected
reduction in posterior uncertainty provided by the annotation, and relies 
on an accurate representation of uncertainty. Details of the
MI estimator and the sequential BED algorithm are in Appendix~\ref{sec:sbed}.

We perform $5$ replicated experiments on the K33 dataset, each with $10$
rounds of annotation. The first round starts with no annotations; successive 
rounds add an annotation. We compare against a baseline that selects 
annotations at random (i.e., planning without a model of uncertainty).

Figure~\ref{fig:k33_uq_reduction_quality_improvement} (left) compares reduction
in posterior trajectory uncertainty between BED annotations (magenta) and
baseline annotations (green). Both posteriors begin with broad uncertainty, but
BED rapidly reduces posterior uncertainty by picking informative annotations
until it stabilizes at a low value by round $5$. In contrast, the baseline
chooses uninformative annotations that don't noticeably reduce uncertainty.

Figure~\ref{fig:k33_uq_reduction_quality_improvement} (right) plots distance to the groundtruth as a
function of annotation round. As above, the informative annotations chosen by
BED (magenta) rapidly improve the quality of posterior samples by reducing their
distance to the groundtruth. In contrast, baseline planning yields little improvement in trajectory quality.

These results show that a small number of automatically-scheduled annotations
enable rapid reduction in posterior uncertainty that correspond to improvements
in track quality. Informative scheduling requires JPT's accurate representation
of uncertainty.


%
\section{Conclusion}
\label{sec:discussion}
We propose JPT, a Bayesian solution to the general multi-object tracking
problem. We construct efficient inference to reason over permutations of
associations and empirically demonstrate that JPT more effectively represents
posterior uncertainty than baselines while outperforming them on standard
tracking metrics. We then show that JPT's accurate representation of uncertainty
enables automatic scheduling of informative disambiguations which rapidly drive
down posterior uncertainty while improving trajectory quality.

\section*{Appendix}

\appendix

\section{Multidimensional Assignment Formulation}
\label{sec:maf}
Multi-object tracking can be formulated in several common ways: as a set
partitioning problem \cite{balas1976set}, a set packing problem
\cite{morefield1977application, zhang2008global}, a maximum-weight independent
set problem \cite{papageorgiou2009maximum} or a multidimensional assignment
problem \cite{poore1994multidimensional}. A review of these formulations is
conducted by \cite{collins2012multitarget}, who shows that the multidimensional
assignment formulation is not limited to pairwise terms as the set-packing,
network-flow solutions \cite{Pirsiavash2011, berclaz2011multiple} are. We next
define the multidimensional assignment problem, as it is most similar to how JPT
is formulated.

Consider $t = 1, \ldots, T$ timesteps with corresponding observation sets $y =
\{ y_1, \ldots, y_T \}$ where the time-$t$ observation set $y_t = \{ y_{tn}
\}_{n=1}^{N_t}$ has $N_t$ observations. Hence, $y_{tn}$ is the $n^{\text{th}}$
observation at time $t$. Let $I_t = \{ 0, 1, \ldots, N_t \}$ be
an index set into $y_t$, where $0$ indicates a false positive or missing
detection. Define $\mathcal{P} = I_1 \times I_2 \times \cdots \times I_T$ as the
set of paths through all index sets such that every path is length-$T$ and has
at least one non-zero index. Interpret a path with a single
non-zero index as a false-positive. Interpret a path with two or more non-zero
indices as an object. Define $\gamma(i_1, \ldots, i_T)$ as a fixed, real cost for path
$(i_1, \ldots, i_T) \in \mathcal{P}$ where $i_t \in I_t$, and $B(i_1, \ldots,
i_T)$ is a boolean variable signifying whether path $(i_1, \ldots, i_T)$ is
included in a solution. Then the multidimensional assignment problem is to find
the $B(i_1, \ldots, i_T)$ that minimizes:
\begin{equation}
\begin{aligned}
  & \min
  && \sum_{i_1=0}^{N_1} 
     \sum_{i_2=0}^{N_2} 
     \ldots
     \sum_{i_T=0}^{N_T} 
     \gamma(i_1, \ldots, i_T)\ B(i_1, \ldots, i_T) 
     \\
  & \text{subject to} && 
  \\
  & && 
  \sum {}
  \sum_{I \setminus i_t} {}
  \ldots
  \sum {}
  B(i_1, \ldots, i_T) = 1
  \qquad
  (\forall i_t = 1, \ldots, N_t, \forall t = 1, \ldots, T)
  \\
  & && 
  B(i_1, \ldots, i_T) \in \{0, 1 \}
  \label{eqn:mot_map}
\end{aligned}
\end{equation}
The objective sums over the costs of all included paths in the solution. For
each observation there is a constraint enforcing that it be claimed by exactly one
path included in the solution (equivalently, that an observation is uniquely
associated either to clutter or a distinct object). MHT \cite{kim2015multiple}
and JPDA \cite{hamid2015joint} are deterministic solutions whereas MCMCDA
\cite{oh2009markov} is a stochastic solution to the multidimensional assignment
problem. The number of possible paths grow exponentially with $T$ and
factorially at each time with $N_t$. Solving this exactly is NP-hard
\cite{blackman1999design, pasiliao2010local}, forcing the above approaches to
use gating heuristics such as a maximum distance between object locations and
observations, a maximum distance between pairwise object locations, or a maximum
number of consecutive missing detections.

We note that JPT represents a departure from the multidimensional assignment
formulation because it does not assign a fixed cost to each association
hypothesis.  While a fixed cost could be constructed--such as by using smoothed
state estimates or marginalizing out all trajectories--our focus is to explore
and represent joint uncertainty in trajectories and data associations.  JPT is
not the only work to depart from a traditional multi-object tracking objective
\cite{andriyenko2011multi}.
\section{JPT Metropolis-Hastings Inference}
We derive the Hastings ratios for the Switch (\ref{sec:switch_ratio}) and Gather
(\ref{sec:gather_ratio}) proposals discussed in the paper. We then treat
trajectory inference when there are missing associations
(\ref{sec:ffbs_missing}).
%
%
\subsection{Switch Proposal Hastings Ratio}
\label{sec:switch_ratio}
We show the derivation of the Switch proposal, recalling that $\sigma_t$ is a
permutation of objects $1, \ldots, K(z)$:
\begin{align}
  R_\text{switch} &=
  \frac{
    p(x', z', M' \mid y)
  }{
    p(x, z, M \mid y)
  }
  \times 
  \frac{
    q_\text{switch}(x, z, M \mid x', z', M', y)
  }{
    q_\text{switch}(x', z', M' \mid x, z, M, y)
  }
  \label{eqn:switch:mh:line1}
  &\\ &=
  \frac{
    p(z' \mid M')\ 
    \prod_{t=1}^T
    \frac{1}{Z}\ 
    p(M_t' \mid M_{t-1}')\ 
    p(x_t' \mid x_{1:t-1}')\ 
    p(y_t \mid x_t', z_t')
  }{
    p(z \mid M)\ 
    \prod_{t=1}^T
    \frac{1}{Z}\ 
    p(M_t \mid M_{t-1})\ 
    p(x_t \mid x_{1:t-1})\ 
    p(y_t \mid x_t, z_t)\ 
  }
  \times
  \label{eqn:switch:mh:line2}
  \\& \qquad \qquad
  \frac{
    \prod_{t=1}^T
    \frac{1}{Z_t}\ 
    p( \sigma_t^{-1}(x_t') \mid \sigma_{1:t-1}^{-1}(x_{1:t-1}')\ 
    p( y_t \mid \sigma_t^{-1}(x_t), \sigma_t^{-1}(z_t) )
  }{
    \prod_{t=1}^T
    \frac{1}{Z_t}\ 
    p( \sigma_t(x_t) \mid \sigma_{1:t-1}(x_{1:t-1})\ 
    p( y_t \mid \sigma_t(x_t), \sigma_t(z_t) )
  }
  \nonumber
  &\\ &=
  \frac{
    p(z' \mid M')\ 
    \prod_{t=1}^T
    p(M_t' \mid M_{t-1}')\ 
    p(x_t' \mid x_{1:t-1}')\ 
    p(y_t \mid x_t', z_t')
  }{
    p(z \mid M)\ 
    \prod_{t=1}^T
    p(M_t \mid M_{t-1})\ 
    p(x_t \mid x_{1:t-1})\ 
    p(y_t \mid x_t, z_t)
  }
  \times
  \label{eqn:switch:mh:line3}
  \\& \qquad \qquad
  \frac{
    \prod_{t=1}^T
    p(x_t \mid x_{1:t-1})\
    p(y_t \mid x_t, z_t)
  }{
    \prod_{t=1}^T
    p(x_t' \mid x_{1:t-1}')\
    p(y_t \mid x_t', z_t')
  }
  \nonumber
  &\\ &=
  \frac{
    \prod_{t=1}^T
    p(M_t' \mid M_{t-1}')
  }{
    \prod_{t=1}^T
    p(M_t \mid M_{t-1})
  }
  \label{eqn:switch:mh:line4}
\end{align}
where Equation~\ref{eqn:switch:mh:line2} substitutes in the values for each term
in the ratio, defining $\sigma_t^{-1}$ as the inverse permutation of $\sigma_t$
and $Z_t$ as the normalizer for the sampled $\sigma_t$ at time $t$ (equal to $1$
if $t \not \in \tau$). Equation~\ref{eqn:switch:mh:line3} substitutes
$\sigma_t(x_t)$ for $x_t'$ and $\sigma_t^{-1}(x_t')$ for $x_t$ (similarly for
$\sigma_t(z_t)$). It also cancels common normalizers $Z$ for the joint ratio and
$Z_t$ at each time $t$ for the proposal ratio.
Equation~\ref{eqn:switch:mh:line4} cancels all terms related to the dynamics and
observation models, and also cancels $p(z' \mid M')$ with $p(z \mid M)$ under
the assumption that no object $k \in \mathcal{K}$ was rendered invalid by having
fewer than two observations. That can easily be detected and automatically
rejected or entirely avoided by defining valid permutations to require the first
two observed times for any $k \in \mathcal{K}$ to not be permutable.
\subsection{Gather Proposal Hastings Ratio}
\label{sec:gather_ratio}
\begin{align}
  \tiny
  R_\text{gather} &=
  \frac{
    p(x', z', M' \mid y)
  }{
    p(x, z, M \mid y)
  }
  \times 
  \frac{
    q_\text{disperse}(x, z, M \mid x', z', M', y)
  }{
    q_\text{gather}(x', z', M' \mid x, z, M, y)
  }
  \label{eqn:gather:mh:line1}
  &\\ &=
  \frac{
    p(z' \mid M')\ 
    \prod_{t=1}^T
    p(M_t' \mid M_{t-1}')\ 
    p(x_t' \mid x_{1:t-1}')\ 
    p(y_t \mid x_t', z_t')
  }{
    p(z \mid M)\ 
    \prod_{t=1}^T
    p(M_t \mid M_{t-1})\ 
    p(x_t \mid x_{1:t-1})\ 
    p(y_t \mid x_t, z_t)\ 
  }
  \times
  \frac{
    (K(z)+1)^{-1}
  }{
    \prod_{t \in \tau_0}
    \omega_t
  }
  \label{eqn:gather:mh:line2}
\end{align}
where $\omega_t = \delta$ if $z_{tn}' \neq k \text{ for any } n$ else $\omega_t
= \frac{1}{Z_t} p(y_{tn} \mid x_{t' k}, z_{tn}' = k)\ p(x_{tk}' \mid x_{t' k}',
z_{tn}' = k) (1-\delta)$.
All dynamics and observation model terms cancel in the posterior ratio for
objects other than $k$, but terms remain for observations that were previously
clutter and are now associated to object $k$ and counts $M' \neq M$.
\subsection{Forward-Filtering, Backward Sampling with Missing Data}
\label{sec:ffbs_missing}
Recall from the main paper that if there are no missing observations (there is
some $n$ at every time $t$ such that $z_{tn}=k$ for each $k \in
\{1, \ldots, K(z)\}$, then the full conditional on trajectories can be sampled
as,
\begin{align}
  p(x \mid z, y)
  &= \prod_{t=1}^T p(x_t \mid x_{t+1:T}, y_{1:T}, z_{1:T}) \\
  &= \prod_{t=1}^T
    \frac{
      p(x_t \mid y_{1:t}, z_{1:t}) \ p( x_{t+1} \mid x_t )
    }{
      p(x_{t+1} \mid y_{1:t}, z_{1:t})
    } \\
  &= \prod_{k=1}^{K(z)} \prod_{t=1}^T
    \frac{
      p(x_{tk} \mid y_{1:t}^k) \ p( x_{(t+1)k} \mid x_{tk} )
    }{
      p(x_{(t+1)k} \mid y_{1:t}^k)
    }
    \label{eqn:ffbs}
\end{align}
where inference is performed independently for each object,
$y_{1:t}^k = \{ y_{t' n} : z_{t' n} = k \text{ and } t' \leq t \}$ and
$p(x_{tk} \mid y_{1:t}^k)$ is the marginal (filter) distribution of $x_{tk}$.

In the case of missing observations, we marginalize over the intervening latent
states, realizing samples only at times where an object has an association.
Thus, the distribution for $x_{tk}$ under the joint (the numerator of
Equation~\ref{eqn:ffbs}), assuming that the most recent previous association
occurred at time $\cev{t} \leq t$ and most recent future association at time
$\vec{t} > t$, is:
\begin{equation}
  p(x_{tk} \mid y_{1:\cev{t}}^k) \ p(x_{\vec{t}k} \mid x_{tk}) = 
  \int
      p( x_{(\cev{t}+1 : t)k} \mid y_{1:\cev{t}}^k)
      d {x_{(\cev{t}+1 : t-1)k}}
  \int
      p(x_{(t+1:\vec{t})k} \mid x_{tk})
      d {x_{(t+1 : \vec{t}-1k)}}.
  \label{eqn:ffbs_missing}
\end{equation}
We emphasize that the first term in Equation \ref{eqn:ffbs_missing} integrates
over past missing states. If there is an association at time $t$ (\ie $\cev{t}
= t$), then there is no integration to carry out in the first term and so it
simplifies to $p(x_{tk} \mid y_{1:t}^k)$. 
Similarly, the second term in Equation
\ref{eqn:ffbs_missing} integrates over future missing states. If $\vec{t} = t+1$
then there is no integration to carry out in the second term and so it
simplifies to $p(x_{(t+1)k} \mid x_{tk})$. Hence, when $\vec{t} = t+1$ and
$\cev{t} = t$, we recover the numerator of Equation~\ref{eqn:ffbs}.
\section{Experiment Details}
We elaborate on experiments in the paper, starting with a description
of each dataset (\ref{sec:datasets}), then detailing the grid search over gating
heuristics we performed to give MCMCDA the best performance
(\ref{sec:mcmcda_grid}) and ending with details on how we compute distances
between sets of associations, used in the uncertainty
quantification and uncertainty reduction experiments of the paper
(\ref{sec:assoc_distance}).

\subsection{Description of Datasets Used in Experiments }
\label{sec:datasets}
K33 is a synthetic dataset containing multiple ambiguous object crossing events.
There are no clutter detections and all objects are detected at all times.

Marmoset contains two primates interacting in a laboratory environment over long
periods of time where there are many total and partial occlusion events, as well
as occasional clutter detections. Noisy observations are generated as the
centroid of the detections from a trained Mask-RCNN neural network
\cite{he2017mask} and groundtruth accomplished by human annotation that
correctly maintains object identities throughout the sequence. As a result, trackers
must correctly re-identify objects that have been occluded to avoid getting
penalized.

Soccer observations are the unassociated centers of players and referee.
Groundtruth does not maintain the identities for objects that go out of frame;
hence, re-identification after a total occlusion is not
rewarded. Metrics are evaluated in chunks of $20$ frames according to the protocol in
\cite{turner2014complete}.
\subsection{MCMCDA Gating Heuristic Grid Search}
\label{sec:mcmcda_grid}
The MCMCDA baseline contains two gating heuristics: thresholds on $\bar{v}$, the
maximum L2 spatial and $d$, the maximum L1 temporal distances between two
observations associated to the same object. Although they can be removed by
setting the thresholds very high, this causes the inference procedures of MCMCDA
to devolve into random exploration, severely damaging performance according to
the CLEAR MOT metrics used in the paper. To make the comparison with JPT as
competitive as possible, we performed a grid search over each gating threshold,
as well as providing it with knowledge of the true number of objects or not, and
restricted JPT/MCMCDA comparisons so that MCMCDA only used the parameters with
best performance as measured by the CLEAR MOT multi-object tracking accuracy
(MOTA) metric.

For MCMCDA on the K33 dataset, the best-performing spatial gating threshold was
$\bar{v}=6$ and the best temporal gating threshold was $d=1$. For Marmoset and
Soccer, the best-performing spatial gating threshold was $\bar{v}=20$ and
best-performing temporal gating threshold was $d = 6$. In all cases, MCMCDA
performed best without knowledge of the true number of objects because this
knowledge could limit its ability to explore by creating excess objects that it
later destroyed. In some cases, it would also cause MCMCDA to be severely
penalized by occlusion events that persisted for longer than its temporal gating
threshold as it could either represent the object before or after the occlusion.

Increasing the gating thresholds to very large numbers caused random exploration
of low-probability events in the MCMCDA posterior due to the way its inference
is constructed.  Specifically, MCMCDA precomputes a sparse graph of paths
between observations that respect its gating thresholds. It then computes
Metropolis-Hastings proposals that randomly sample from this graph on the
assumption that the thresholds were set to encourage likely associations. Thus,
MCMCDA inference has a fundamental limitation: either gating thresholds are set
tight and some true association hypotheses are 
excluded, or they are set loose and random exploration occurs.
\subsection{Computing Distances Between Association Hypotheses}
\label{sec:assoc_distance}
To match an inferred set of trajectories to another set of trajectories, we
begin with the Spatiotemporal Linear Combine (STLC) Distance of
\cite{shang2017trajectory}, which compared favorably in \cite{su2019survey}.
Briefly, STLC evaluates trajectories on both their L2 spatial and L1 temporal
alignment; it supports uneven sampling rates and arbitrary trajectory start/end
times. It is a similarity measure that ranges from $[0, 2]$, but we convert it
to a cost by inverting the limits.

Given STLC as an object-to-object cost, we define a distance between
multi-object tracking association hypotheses by using discrete optimal transport
\cite{solomon2018optimal}, where the cost matrix is filled with the STLC costs
of each object pair between the two samples. Note that this supports arbitrary
numbers of objects in each sample. This distance was then used in determining
mode representation in posterior samples of JPT and MCMCDA in the Uncertainty
Quantification experiments and again in the Uncertainty Reduction experiments,
where we demonstrated that planned annotations rapidly reduce the distance of
JPT samples to the groundtruth.
\section{Sequential Bayesian Experiment Design}
\label{sec:sbed}
Bayesian experiment design (BED) optimizes a utility function over 
the set of observation pairs for annotation, where each observation pair
$(y_{t_1 n_1}, y_{t_2 n_2})$ is termed a \emph{design}.
The sequential form of BED allows annotation results from earlier rounds of BED 
to inform the selection of designs in subsequent rounds.
Mutual information (MI) is a commonly used utility function for BED 
that quantifies the expected reduction in posterior uncertainty that results from 
annotation of a design. 
MI is especially suited for our task since we seek to reduce uncertainty in the 
trajectory posterior and has several appealing properties include invariance to 
reparameterization. We next describe the annotation model in detail.

Let $\kappa = (t, n)$ be the time and observation indices that uniquely 
identifies observation $y_{tn}$. A design then corresponds to a tuple of these index pairs
$d = (\kappa_1, \kappa_2)$. We abuse notation and let $\kappa_1(d) = (t_1, n_1)$ 
indicate the first pair in design $d$ 
such that $y_{\kappa_1(d)} = y_{t_1 n_1}$ and $z_{\kappa_1(d)} = z_{t_1 n_1}$ 
and likewise for $\kappa_2(d)$.
Recall from the main text that the annotation 
indicates if two observations $y_{t_1 n_1}, y_{t_2 n_2}$ belong to the 
same or different objects -- $a_l(y_{t_1 n_1}, y_{t_2 n_2}) =1 \text{ or } 0$ 
respectively -- and is correct with probability $p_a = 0.99$. This event corresponds to 
whether the assignments $z_{t_1 n_1}, z_{t_2 n_2}$ share the same non-zero value
-- recall $z_{tn} = 0$ indicates clutter.
After accounting for the annotation noise and design, we have the following annotation likelihood
\begin{equation}
p_d(a_l = 1 \mid x, y, z, M) = p_d(a_l = 1 \mid z_{\kappa_1(d)}, z_{\kappa_2(d)})=
\begin{cases}
  0.99 
   &\text{ if  } z_{\kappa_1(d)} = z_{\kappa_2(d)} \text{ and } z_{\kappa_1(d)} > 0 \\
  0.01
   &\text{ o.w. }\label{eqn:pa}
  \end{cases}
\end{equation}
When conditioned on just the two assignments $z_{\kappa_1(d)}, z_{\kappa_2(d)}$, 
the annotation is independent of the 
remaining variables in the model; this yields the first equality
in Equation \ref{eqn:pa}. This conditional taken in conjunction with the 
joint distribution $p(x, y, z, M)$ described in the paper yields a complete
generative model that now includes annotations.

Mutual information between the annotation $a_l$ and the latent trajectories $x$
conditioned on the observations $y$ and past annotations 
$D=\{a_{1:l-1}, d_{1:l-1}\}$ is given by,
\begin{align}
I_d(a_l;x \mid y, D ) 
  &= \mathbb{E}\left[
  \log \frac{p_d(a_l, x \mid y, D )}{p_d(a_l \mid y, D )p_d(x \mid y, D )} \right]\\
  &= \mathbb{E}\left[
  \log \frac{p_d(a_l \mid x, y, D )}{p_d(a_l \mid y, D )} \right]\\
  &= \mathbb{E}\left[- \log p_d(a_l \mid y, D )\right] 
  - \mathbb{E}\left[- \log p_d(a_l \mid x, y, D ) \right].\label{eqn:entropy}
\end{align}
We highlight that Equation \ref{eqn:entropy} is the difference of entropies
(a measure of uncertainty for random variables), 
illustrating how MI is a measure of the expected reduction in uncertainty.

We use a greedy approach to Sequential BED, wherein we select the 
highest MI design within each round of BED. While myopic, this approach avoids 
the complexity associated with searching for an optimal policy. Thus, at the 
$l^\text{th}-$round of sequential BED, we seek 
\begin{equation}
d_l = \arg\max_d I_d(a_l; x \mid y, D).
\end{equation}

We typically cannot evaluate MI in closed form and instead resort 
to Monte Carlo estimation using $M$ samples drawn from the posterior 
$\{a_{l}^m, x^m, z^m\}_{m=1}^M \sim p(a_l, x, z \mid y, D)$: 
\begin{align}
\tilde{I}_d 
= \frac{1}{M} 
\sum_{m = 1}^M \log 
  \frac
    {p_d(a_l^m \mid x^m, y, D)}
    {p_d(a_l^m \mid y, D)}. \label{eqn:mi_estim}
\end{align} 
To evaluate the likelihoods in Equation \ref{eqn:mi_estim}, 
first we expand them as,
\begin{align}
p_d(a_l^m \mid x^m, y, D ) 
&= \sum_z p_d(a_l^m \mid z, x^m, y, D)\ p(z \mid x^m, y, D) \\
&= \sum_{z_{\kappa_1(d)}, z_{\kappa_2(d)}} p_d(a_l^m \mid z_{\kappa_1(d)}, z_{\kappa_2(d)})\ p(z_{\kappa_1(d)}, z_{\kappa_2(d)} \mid x^m, y, D).\label{eqn:pa_posterior}\\
p_d(a_l^m \mid y, D) 
&= \sum_z p_d(a_l^m \mid z, y, D)\ p(z \mid y, D) \\
&= \sum_{z_{\kappa_1(d)}, z_{\kappa_2(d)}} p_d(a_l^m \mid z_{\kappa_1(d)}, z_{\kappa_2(d)})\ p(z_{\kappa_1(d)}, z_{\kappa_2(d)} \mid y, D) \label{eqn:pa_prior} 
\end{align}
Equation \ref{eqn:pa_posterior} can be evaluated exactly because we 
can obtain $p(z_{\kappa_1(d)}, z_{\kappa_2(d)} \mid x^m, y, D)$ through 
enumeration of all pairwise assignments conditioned on the sampled trajectories 
$x^m$ and observations $y$. Equation \ref{eqn:pa_prior}, on the other hand, requires 
$p_d(a_l^m \mid y, D)$ which is intractable, so we again 
use Monte Carlo estimation, 
\begin{align}
\hat{p}_d(a_l^m \mid y, D) 
&= \frac{1}{M} \sum_{m' = 1}^M  p_d(a_l^m \mid z_{\kappa_1(d)}^{m'}, z_{\kappa_2(d)}^{m'}).\label{eqn:pa_prior_estim} 
\end{align}
Our MI estimator is then,
\begin{align}
\hat{I}_d 
= \frac{1}{M} 
\sum_{m = 1}^M \log 
  \frac
    {p_d(a_l^m \mid x^m, y, D)}
    {\hat{p}_d(a_l^m \mid y, D)} \label{eqn:mi_estim_final}
\end{align} 
where $p_d(a_l^m \mid x^m, y, D)$ is given in Equation \ref{eqn:pa_posterior} 
and $\hat{p}_d(a_l^m \mid y, D)$ is given in Equation \ref{eqn:pa_prior_estim}.
\section{Switch Proposals Generalize Extended HMM Proposals}
\label{sec:switch_ehmm}
\begin{figure}[]
  \centering
  \includegraphics[width=0.35\textwidth]{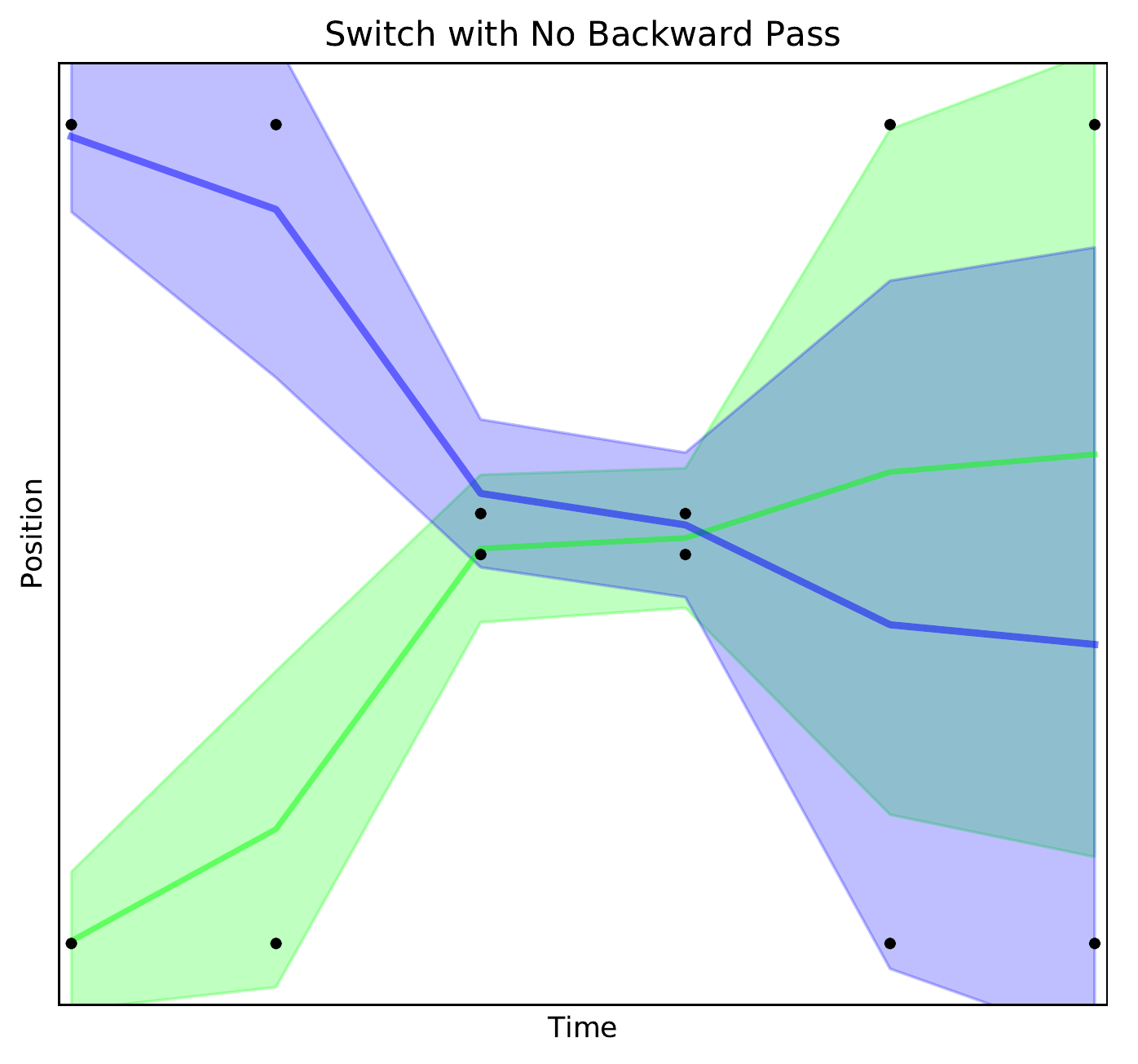}
  \hspace{1cm}
  \includegraphics[width=0.35\textwidth]{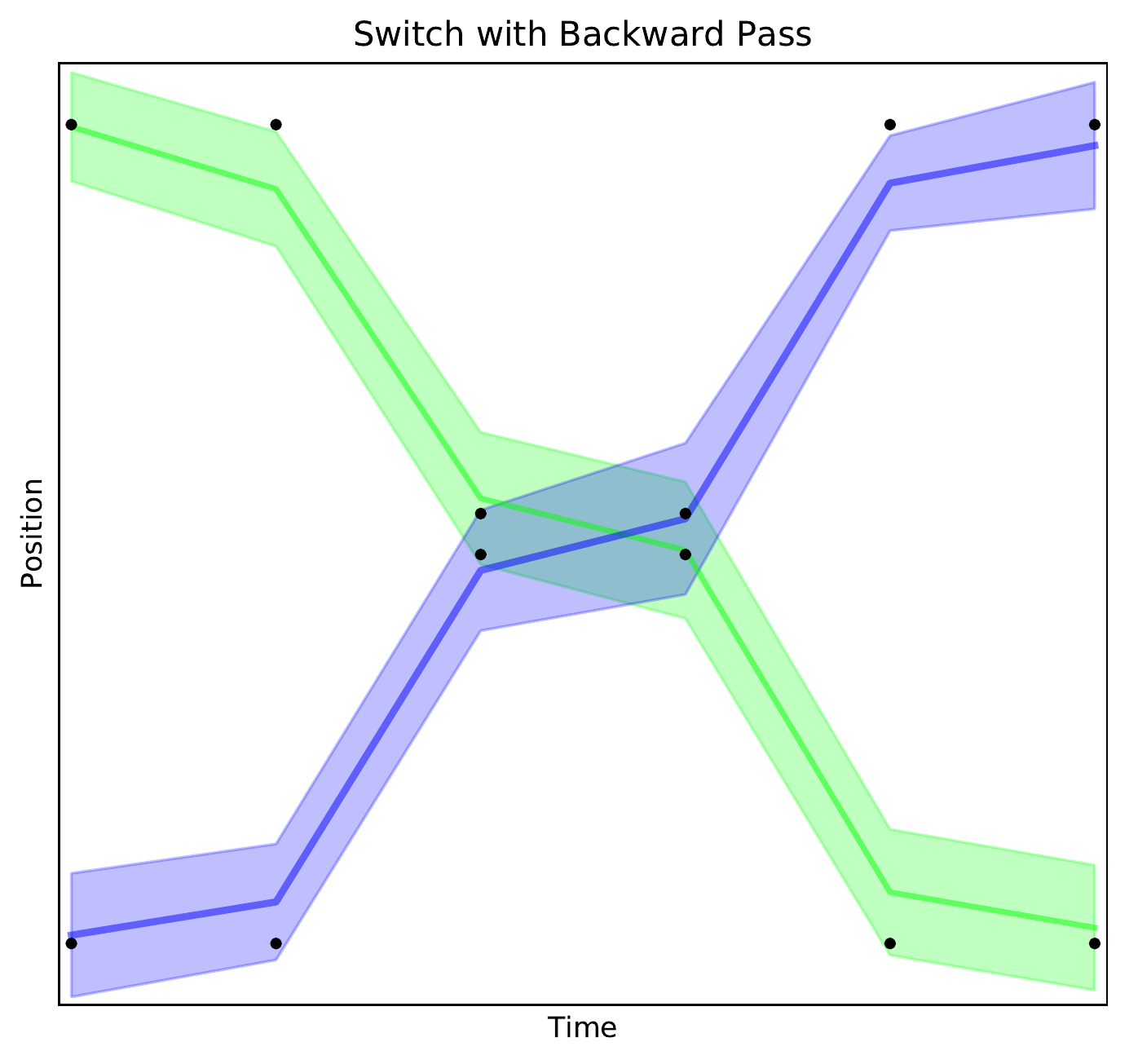}
  \caption{Observations $y$ (black points) with two modes (blue, green objects
  crossed or not). Shading indicates marginal posterior trajectory variance.
  (\textbf{Left}): Switch statements that leave no future associations fixed
  have strong ability to explore modes because future associations do not force
  an outcome.  (\textbf{Right}): Switch statements that leave future
  associations fixed (from the final timestep) will favor the modes supported by
  those fixed associations.}
  \label{fig:switch_forward_backward}
\end{figure}
Switch proposals generalize the Extended HMM (EHMM) proposals of
\cite{neal2004inferring} by permitting discretizations that depend on the
latent space. In brief, EHMM proposals compose an inference method that helps
explore a posterior distribution by proposing a discretization of latent states
(called "pool states") over time. A hidden Markov model is then defined over the
pool states and a joint sample drawn using forward-filtering, backward sampling.
Crucially, the discretization sampled by an EHMM proposal includes the current
latent state, but \emph{must not} otherwise depend on it. If it does, detailed
balance is lost because calculating the reverse move probability would require a
difficult integration over the latent space.

In contrast, Switch proposals sample from a discretization that depends on the
current latent state while maintaining detailed balance. In the nomenclature of
EHMM proposals, the "pool states" of JPT's Switch proposal are permutations of
latent state $x, z$. JPT then samples from the generative model of an HMM
that contains no future information; thus, no backwards pass is required. We
note that the Switch proposal always contains the current state as represented
by the identity permutation over all times.

Switch proposals can easily be constructed to contain future information by
restricting switch times $\tau$ so that some future associations from the sample
set $\mathcal{K}$ remain fixed. Doing so causes a need for future information to
propagate backward. We found that doing so without being careful about which
associations to leave fixed impairs the ability of Switch proposals to
explore different modes as future information encouraged the current sample to
remain in the same mode. See Figure~\ref{fig:switch_forward_backward} for an
example. Backward propagation \emph{is} desirable when future information
comes from annotations since they are intended to reduce posterior uncertainty.
But, in the absence of annotations, future information in the form of
restricted Switch times is not desirable.



\maketitle

\bibliography{main}

\end{document}